\documentclass[runningheads]{llncs}

\usepackage{eccv}

\usepackage{eccvabbrv}

\usepackage{graphicx}
\usepackage{booktabs}
\usepackage[table]{xcolor}
\usepackage[accsupp]{axessibility}

\usepackage[breaklinks,colorlinks,citecolor=eccvblue]{hyperref}

\usepackage{orcidlink}
\usepackage{fontawesome}
\usepackage{cuted}
\usepackage{animate}
\usepackage{caption}
\usepackage{float}
\usepackage{xcolor}
\usepackage{subcaption}
\usepackage{tabularx}
\usepackage{capt-of}
\usepackage{pifont}
\usepackage{booktabs}
\usepackage{multirow}
\usepackage{graphicx}
\usepackage{siunitx}
\usepackage{soul}

\newcommand{\myparagraph}[1]{\vspace{4pt}\noindent\textbf{#1}}
\newcommand{\equalcontrib}{\textsuperscript{$\ddagger$}}

\newcommand{\maketitlesupplementary}{%
  \clearpage
  \appendix
  \setcounter{section}{0}
  \setcounter{subsection}{0}
  \renewcommand{\thesection}{\Alph{section}}%
  \renewcommand{\thesubsection}{\Alph{section}.\arabic{subsection}}%
  \renewcommand{\theHsection}{supp.\Alph{section}}%
  \renewcommand{\theHsubsection}{supp.\Alph{section}.\arabic{subsection}}%
  \begin{center}
    {\Large\bfseries Supplementary Material}
  \end{center}
  \vspace{1.5em}
}

\begin{document}

\title{AirZoo: A Unified Large-Scale Dataset for Grounding Aerial Geometric 3D Vision}

\titlerunning{AirZoo}

\author{Xiaoya Cheng\orcidlink{0000-0002-9793-5640}\inst{1}\equalcontrib \and
Rouwan Wu\orcidlink{0009-0007-1018-8106}\inst{1}\equalcontrib \and
Xinyi Liu\orcidlink{0009-0001-9902-8622}\inst{1}\equalcontrib \and
Zeyu Cui\orcidlink{0009-0002-3639-944X}\inst{1}\equalcontrib \and
Yan Liu\orcidlink{0000-0002-2015-4301}\inst{2}\equalcontrib \\[4pt]
Na Zhao\orcidlink{0000-0003-2329-7014}\inst{3} \and
Yu Liu\orcidlink{0000-0002-3914-1252}\inst{1} \and
Maojun Zhang\orcidlink{0000-0001-6748-0545}\inst{1} \and
Shen Yan\orcidlink{0000-0002-1415-5113}\inst{1}\thanks{Corresponding author.}}

\authorrunning{X.~Cheng et al.}

\institute{National University of Defense Technology, Changsha, China\\
\email{\{chengxy, wurouwan97, liuxinyi24, czy806, jasonyuliu, mjzhang, yanshen12\}@nudt.edu.cn}
\and
National Key Laboratory of Advanced Guidance and Control Technology, Changsha, China\\
\email{cowisee633@gmail.com}
\and
Singapore University of Technology and Design, Singapore\\
\email{na\_zhao@sutd.edu.sg}}

\makeatletter
\apptocmd{\@maketitle}{
   {\centering\small $\ddagger$ Equal contribution.\par}
   \centering
   {\small\url{https://nudt-sawlab.github.io/AirZoo/}}\\[6pt]
   \includegraphics[width=\textwidth]{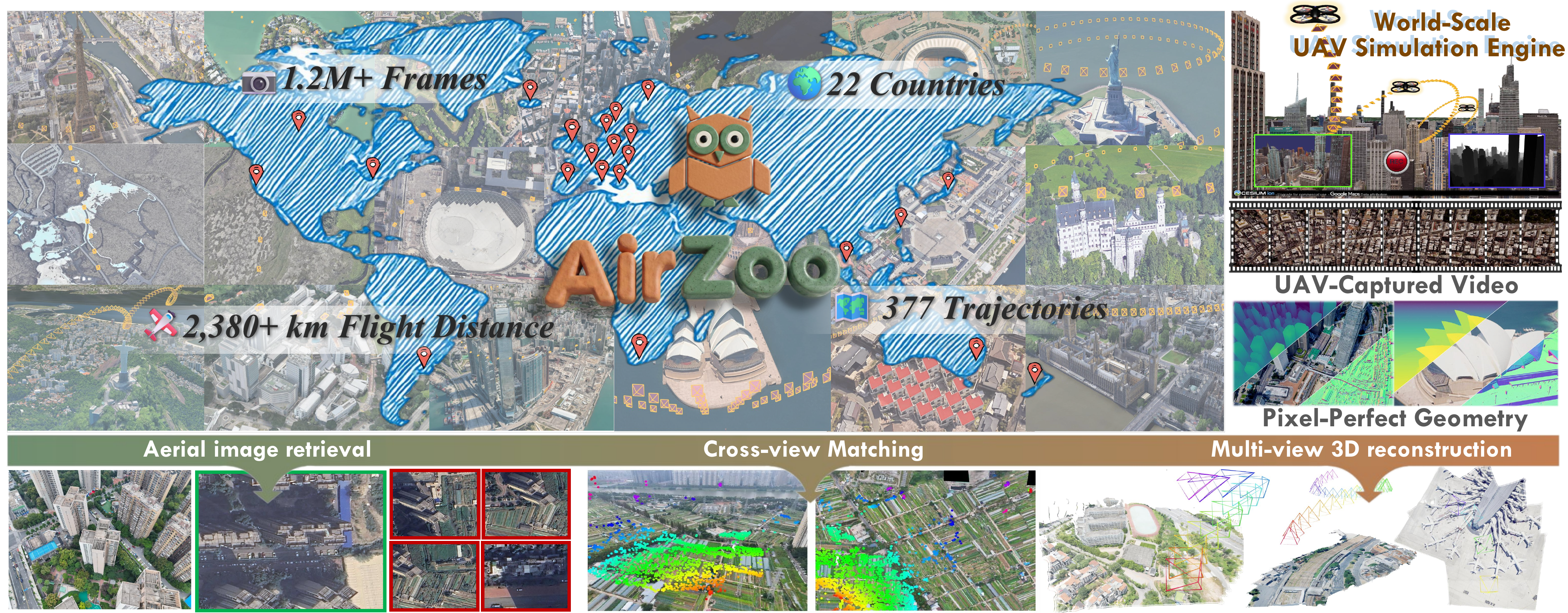}
   \captionof{figure}{\textbf{Overview of AirZoo.} (Top) A million-scale synthetic UAV dataset with pixel-perfect geometry featuring extensive flight trajectories across diverse global regions. (Bottom) Fine-tuning on AirZoo empowers SoTA models to achieve superior performance in aerial geometric 3D vision tasks, including \textbf{\textcolor[HTML]{4F6335}{Aerial image Retrieval}}, \textbf{\textcolor[HTML]{6B582C}{Cross-view Matching}}, and \textbf{\textcolor[HTML]{915020}{Multi-view 3D Reconstruction}}.}
\label{fig:teaser}
}{}{}
\makeatother

\maketitle

\begin{abstract}
Despite the rapid progress in data-driven 3D vision, aerial geometric 3D vision remains a formidable challenge due to the severe scarcity of large-scale, high-fidelity training data. 
Existing benchmarks, predominantly biased toward ground-level or object-centric views, do not account for complex viewpoint transformations and diverse environmental conditions in UAV-based sensing.
    To bridge this critical gap, we propose \textbf{AirZoo}, a unified large-scale dataset and benchmark for grounding aerial geometric 3D vision. AirZoo possesses three appealing properties: \textbf{1) Scalable Generation Pipeline:} Leveraging freely available, world-scale photogrammetric 3D meshes, it renders vast outdoor environments with customizable UAV flight trajectories and configurable weather/illumination. \textbf{2) Comprehensive Scene Diversity:} It provides extensive global coverage (22 countries, 95 base flight sequences, and 377 weather-conditioned trajectories), systematically encompassing both highly structured urban landscapes and complex unstructured natural environments. \textbf{3) Rich Geometric Annotations:} Each frame provides synchronized, pixel-level metric depth and precise 6-DoF geo-referenced poses, essential for geometry-aware learning.
Through three rigorous evaluation tracks—aerial image retrieval, cross-view matching, and multi-view 3D reconstruction—we demonstrate that AirZoo serves as a powerful "pre-training engine." Extensive experiments on both public and newly collected real-world benchmarks reveal that fine-tuning on AirZoo yields substantial performance gains for state-of-the-art models (e.g., MegaLoc, RoMa, VGGT, and Depth Anything 3), establishing a new performance upper bound for aerial spatial intelligence.
\end{abstract}


\section{Introduction}
\label{sec:intro}

Geometric 3D vision, encompassing fundamental tasks such as image-based retrieval, feature matching, and multi-view reconstruction, serves as the cornerstone of spatial intelligence and is indispensable for autonomous robotics and mixed reality.
Recently, the advent of foundation models with large-scale dataset training has yielded unprecedented performance milestones. However, these advancements are largely tethered to a terrestrial-centric bias, as mainstream training corpora are predominantly harvested from ground-level~\cite{Geiger2013IJRR,li2018megadepth,dai2017scannet,yeshwanth2023scannet++,sun2020scalability,loiseau2025rubik} or object-centric~\cite{liu2022akb,fu20213d,chang2015shapenet,Wu_2023_CVPR} perspectives.
This inherent bias creates a severe domain gap when these models are deployed on UAV platforms. Unlike ground-based conditions, UAVs operate in a highly unconstrained 6-DoF space, introducing unique geometric challenges: drastic viewpoint shifts from oblique to nadir, the frequent replacement of horizontal facades with roofs, and extreme scale variations due to changing flight altitudes. Consequently, there is an urgent imperative to curate a large-scale, geometry-aware aerial dataset for these foundation models to unlock the potential of aerial spatial intelligence.

Existing UAV datasets struggle to simultaneously deliver the global scale and dense geometric supervision required for robust 3D vision (see \cref{tab:uav-dataset-comparison}). While many existing UAV datasets offer photorealism~\cite{zhu2023sues, dai2023vision, xu2024uav,vuong2025aerialmegadepth}, they often lack the scale and complete geometric ground truth (e.g., 6-DoF poses, metric depth) needed for robust training. Even recent efforts providing geometric data~\cite{wu2024uavd4l, dhaouadi2025ortholoc, li2023matrixcity, wang2025uavscenes, ji2025game4loc,fonder2019mid,rizzoli2023syndrone,gross2025occufly} are limited to a few manually crafted city models and lack viewpoint variety, making them insufficient for sequential geometric supervision. Alternatively, while some works use freely available, global-scale 3D models like Google Maps~\cite{zheng2020university, berton2024meshvpr} to achieve unbounded spatial coverage, they are mainly designed for image retrieval tasks. Consequently, they only provide photometric data, lacking the underlying 3D geometric ground truth entirely.

To boost the research on aerial geometric 3D vision, we present \textbf{AirZoo}, a large-scale synthetic dataset designed to bridge the critical ground-to-aerial data gap. Our dataset has several appealing properties: \textbf{1) Scalable Generation Pipeline:} We develop a fully automated simulator based on a custom AirSim-Cesium-Unreal Engine pipeline. The pipeline can automatically generate diverse environmental conditions, including varying weather and lighting, ensuring high-fidelity rendering that reflects the complexity of the real world. \textbf{2) Comprehensive Scene Diversity:} Unlike previous works limited to closed worlds, our collection spans 22 countries with 95 base flight sequences rendered into 377 weather-conditioned trajectories. This massive scale covers a wide range of operational UAV environments, ranging from dense urban centers to rural landscapes, to ensure robust generalization. \textbf{3) Rich Geometric Annotations:} Each flight sequence is captured with synchronized sensors, providing RGB images, metric depth maps, and absolute geo-coordinates. This compels the network to learn features grounded in stable 3D geometry rather than transient textures.

To systematically evaluate the utility of AirZoo, we propose a comprehensive evaluation framework spanning three core aerial vision tasks: a) aerial image retrieval, b) cross-view matching, and c) multi-view 3D reconstruction. 
Beyond standard public benchmarks, we introduce AirZoo-Real, a real-world dataset designed to test the limits of current algorithms. By fine-tuning a suite of foundation models—including MegaLoc~\cite{berton2025megaloc}, RoMa~\cite{edstedt2024roma}, VGGT~\cite{wang2025vggt}, and Depth Anything 3~\cite{lin2025depth}—on the synthetic diversity of AirZoo, we demonstrate substantial performance gains across all tracks. Most notably, our results reveal that AirZoo imparts critical domain-invariant geometric knowledge, empowering these models with robust zero-shot generalization capabilities when deployed in complex, unseen real-world UAV scenarios.

Our main contributions are summarized as follows:
\begin{itemize}
    \item We introduce AirZoo, a million-scale synthetic UAV dataset covering diverse global regions with pixel-perfect geometric supervision (metric depth, verified poses).
    \item We propose a scalable AirSim-Cesium-Unreal simulator that enables automated, high-fidelity rendering of global 3D tiles with varying weather and lighting.
    \item We finetune SoTA methods (MegaLoc, RoMa, VGGT, Depth Anything 3) on aerial retrieval, matching, and reconstruction tasks, demonstrating that training on AirZoo significantly improves the performance.
\end{itemize}

\section{Related Work}
\label{sec:related-work}
\begin{table*}[t]
\centering
\caption{
Comparison with large-scale UAV visual geometry datasets. 
\textbf{Regions} denotes geographically distinct environments. (Note that University-1652 and SUES-200 are limited to isolated landmarks, while AirZoo covers city-scale regions).
\textbf{Condition} refers to environmental variations. \textbf{Sequence} denotes continuous flight trajectories. 
\textbf{Pose} indicates precise 6-DoF trajectory. 
\ding{51} indicates available, \textcolor{gray}{\ding{55}} unavailable. 
}
\small
\setlength{\tabcolsep}{4pt} 

\resizebox{\linewidth}{!}{
    \begin{tabular}{lccccccccc}
    \toprule
    \textbf{Dataset} & \textbf{Images} & \textbf{Regions}  & \textbf{Altitude (m)} & \textbf{Pitch} & \textbf{Type} & \textbf{Condition} & \textbf{Pose} & \textbf{Depth} & \textbf{Sequence} \\
    \midrule
    University-1652~\cite{zheng2020university} & 37.8k & 1652 & - & Varied & Syn. & \textcolor{gray}{\ding{55}}  & \textcolor{gray}{\ding{55}} & \textcolor{gray}{\ding{55}}& \textcolor{gray}{\ding{55}} \\
    SUES-200~\cite{zhu2023sues} & 24.1k & 200 & 150--300 & Varied & Real & \textcolor{gray}{\ding{55}} & \textcolor{gray}{\ding{55}} &\textcolor{gray}{\ding{55}} & \ding{51}\\
    DenseUAV~\cite{dai2023vision} & 27k & 14 & 80--100 & Fixed & Real & \ding{51} & \textcolor{gray}{\ding{55}}& \textcolor{gray}{\ding{55}} & \textcolor{gray}{\ding{55}} \\
    UAV-VisLoc~\cite{xu2024uav} & 6.7k & 11 & 400--2k & Fixed & Real & \ding{51} & \textcolor{gray}{\ding{55}} & \textcolor{gray}{\ding{55}}& \ding{51} \\
    UAVD4L~\cite{wu2024uavd4l} & 6.8k & 1 & 50--300 & Varied & Real & \textcolor{gray}{\ding{55}} & \ding{51}& \ding{51}& \textcolor{gray}{\ding{55}} \\
    Game4Loc~\cite{ji2025game4loc} & 33.7k & 1 & 80--650 & Varied & Syn. & \textcolor{gray}{\ding{55}} & \ding{51}& \textcolor{gray}{\ding{55}} & \textcolor{gray}{\ding{55}}\\
    MatrixCity~\cite{li2023matrixcity} & 67k & 2 & 100--450 & Oblique & Syn. & \ding{51}& \ding{51}& \ding{51} & \ding{55} \\
    UAVScenes~\cite{wang2025uavscenes} & 120k & 4 & 80--130 & Fixed & Real & \textcolor{gray}{\ding{55}} & \ding{51}& \ding{51}& \ding{51} \\
    OrthoLoC~\cite{dhaouadi2025ortholoc} & 16.4k & 47 & 50--300 & Varied & Real & \textcolor{gray}{\ding{55}} & \ding{51}& \ding{51}& \textcolor{gray}{\ding{55}} \\
    \midrule
    \textbf{AirZoo (Syn)} & \textbf{1.2M} & \textbf{95} & \textbf{0--800} & \textbf{Varied} & \textbf{Syn.} & \textbf{\ding{51}} & \textbf{\ding{51}} & \textbf{\ding{51}} & \textbf{\ding{51}}\\
    \textbf{AirZoo (Real)} & \textbf{9.4k+} & \textbf{2} & \textbf{0--350} & \textbf{Varied} & \textbf{Real} & \textbf{\ding{51}} & \textbf{\ding{51}} & \textbf{\ding{51}}& \textbf{\ding{51}} \\
    \bottomrule 
    \end{tabular}
}

\label{tab:uav-dataset-comparison}
\end{table*}

\noindent\textbf{UAV-Specific Datasets.} Constructing a large-scale aerial benchmark with precise geometry is non-trivial. Existing datasets struggle to simultaneously balance global scale, environmental diversity, and dense geometric supervision. Specifically, expansive aerial collections and large-scale retrieval benchmarks~\cite{zhu2023sues, dai2023vision, xu2024uav, vuong2025aerialmegadepth, zheng2020university, berton2024meshvpr} offer high photorealism or expansive coverage, but they fundamentally lack complete geometric ground truth. To circumvent this, geometry-centric datasets~\cite{wu2024uavd4l, dhaouadi2025ortholoc, li2023matrixcity, wang2025uavscenes, ji2025game4loc, fonder2019mid, rizzoli2023syndrone, gross2025occufly} provide dense annotations, yet they are severely restricted to a few localized scenes or specific city models with limited viewpoint variety. To bridge this critical gap, we propose AirZoo to unlock the geometric potential of world-scale 3D maps. By introducing an automated pipeline that simulates realistic UAV flight trajectories to render continuous sequences from global 3D tiles, AirZoo uniquely pairs immense geographic variety with synchronized dense geometry. This sequence-level design provides the essential spatial context required for robust 3D vision, significantly boosting zero-shot generalization on real-world aerial tasks.

\noindent\textbf{Aerial Image Retrieval.}
Cross-view image retrieval establishes the similarity between UAV and satellite imagery to achieve coarse UAV localization. Due to the high cost of collecting real-world data, University-1652~\cite{zheng2020university} utilized Google Earth to simulate circle flights, constructing the first cross-view UAV-satellite dataset and employing contrastive learning for discriminative feature learning. With the advent of foundation models~\cite{oquab2023dinov2}, AnyLoc~\cite{keetha2023anyloc} combined large-scale pre-trained models with training-free aggregation, proposing a unified method for various sub-tasks. However, given the scarcity of cross-view data and ground truth, most current approaches still adopt methodologies from ground-based place recognition, such as SALAD~\cite{izquierdo2024optimal} and MegaLoc~\cite{berton2025megaloc}, which benefit from more extensive training data.

\noindent\textbf{Cross-view Matching.}
Cross-view matching aims to establish dense correspondences or feature associations between images captured from drastically different viewpoints. Previous works have largely focused on the ground-to-aerial domain, utilizing datasets such as AerialMegadepth~\cite{vuong2025aerialmegadepth} and BlendedMVS~\cite{yao2020blendedmvs}. However, state-of-the-art matching~\cite{edstedt2024roma, sun2021loftr} methods are predominantly trained on ground-to-ground datasets like MegaDepth. Consequently, these models suffer from a significant domain gap when applied to aerial-to-satellite matching. Due to the lack of dedicated training on top-down satellite imagery, their capability to handle the extreme geometric distortions and scale variations in aerial-satellite pairs is severely limited.

\noindent\textbf{Multi-view visual geometry estimation.} Traditional pipelines and early learning methods~\cite{schonberger2016structure,schonberger2016pixelwise,kendall2015posenet,sarlin2021back,li2020hierarchical,brachmann2023accelerated,wang2024glace} heavily rely on stable correspondences and struggle in ill-posed conditions. Recently, feed-forward Transformers~\cite{wang2024dust3r,wang2025vggt,yang2025fast3r,cabon2025must3r,zhang2024monst3r,lin2025depth} shifted the paradigm by directly predicting point maps to jointly recover depth and poses. Leveraging massive training data and unified architectures, these methods have achieved state-of-the-art performance and remarkable generalization across most standard scenarios. However, since their training datasets mainly consist of ground-level images~\cite{Geiger2013IJRR,li2018megadepth,dai2017scannet,yeshwanth2023scannet++,sun2020scalability,loiseau2025rubik}, deploying these general-purpose models on UAVs reveals a severe domain gap. They struggle with unique aerial characteristics such as drastic oblique-to-nadir viewpoint shifts and unstructured terrains. Consequently, scalable and robust geometry estimation tailored for UAVs remains an open challenge. In this work, we leverage our constructed dataset to empower these foundation models, effectively unlocking their potential for aerial applications.

\section{The AirZoo Dataset}
\label{sec:dataset}
\begin{figure}[t]
  \centering
  \includegraphics[width=\linewidth]{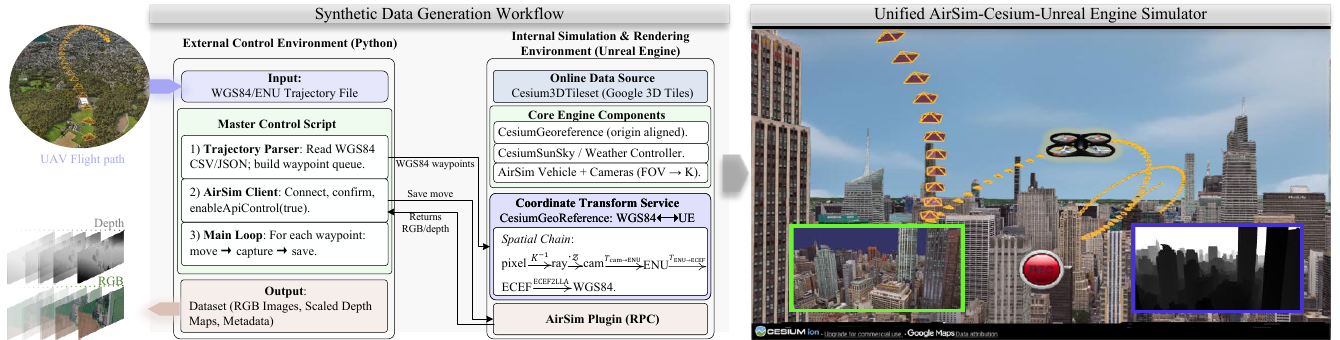}
  \caption{\textbf{Overview of the AirSim-Cesium-Unreal simulator.} Python issues WGS84 trajectories, Cesium for Unreal streams Google 3D Tiles into UE, and AirSim controls synchronized RGB, depth, and georeferenced pose capture.}
  \label{fig:pipeline_workflow}
\end{figure}

In this section, we describe the data generation framework and the statistics of AirZoo.

\subsection{Data Generation and Processing}

\noindent\textbf{Global Region Selection.}
We utilize Cesium for Unreal, a plugin that streams
Google 3D Tiles directly into Unreal Engine 5 (UE5). These tiles provide the
highest level of detail for global-scale 3D photogrammetry models currently available. We select regions across six continents, covering urban, rural, and natural terrains for subsequent UAV trajectory simulations. Together, these curated
regions establish the bases for our subsequent UAV trajectory simulations.

\noindent\textbf{Automated Rendering Pipeline.}
We develop a custom AirSim-Cesium-Unreal simulator for continuous UAV trajectories over global terrains (Fig.~\ref{fig:pipeline_workflow}). The system integrates three core components:
\begin{itemize}
    \item \textbf{Unreal Engine} (UE) provides physically based rendering and supports dynamic weather and illumination.
    \item \textbf{Cesium for Unreal} streams Google 3D Tiles into UE and anchors the simulation world to geographic coordinates via Cesium Georeference.
    \item \textbf{AirSim} simulates the multi-rotor UAV and virtual cameras, executing WGS84 waypoints with reproducible pose control.
\end{itemize}
A Python controller orchestrates the acquisition loop across these components. It parses predefined trajectories, sends waypoint commands through the AirSim RPC interface, and performs on-the-fly WGS84-to-UE coordinate conversion. For each waypoint, AirSim's Steppable Clock coordinates tile loading and sensor capture, allowing the simulator to move the UAV to the target pose and record $\mathcal{M}=\{\textit{RGB}, \textit{depth}, \textit{intrinsics}, \textit{pose}\}$ in a synchronized step after the streamed tiles are stabilized. This controlled loop provides deterministic captures with reliable geometric annotations for downstream learning.

\begin{figure*}[t]
  \centering
  \includegraphics[width=\linewidth]{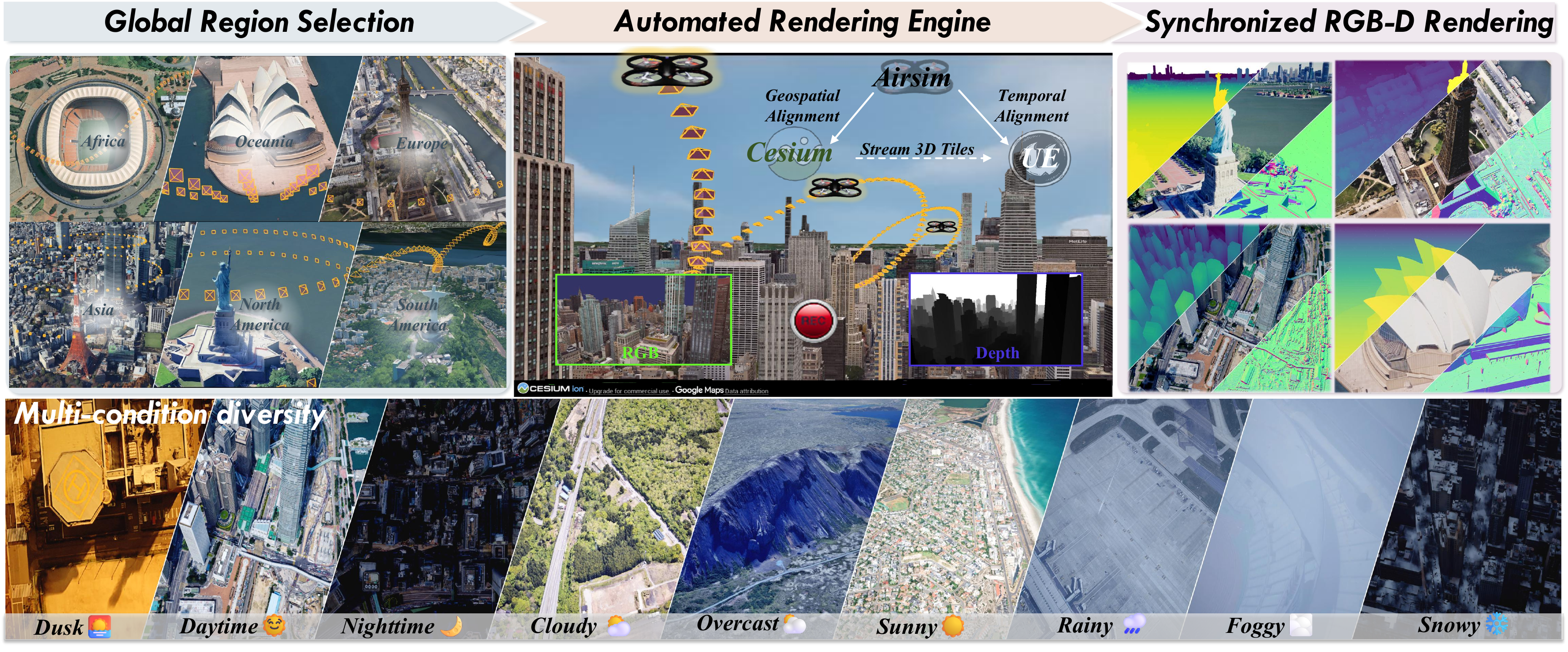}
  \caption{\textbf{Construction pipeline and properties of AirZoo.} Global region selection (top-left), simulation (top-middle), synchronized RGB-D rendering (top-right), and multi-condition diversity (bottom).}
  \label{fig:system_overview_spacious}
\end{figure*}

\noindent\textbf{Synchronized RGB-D Rendering.}
Driven by the aforementioned pipeline, we record continuous 30\,fps flight trajectories rather than isolated keyframes, generating temporally consistent data along each trajectory.  As illustrated in Fig.~\ref{fig:system_overview_spacious} (top-right), each rendered frame provides an RGB image coupled with pixel-aligned absolute depth (in meters) and calibrated camera intrinsics. By extracting these complete modalities, our engine overcomes the limitations of previous works~\cite{zheng2020university, berton2024meshvpr} that rely solely on photometric screenshots, enabling true geometry-aware learning. Furthermore, we leverage UE5's dynamic volumetric systems to simulate diverse environmental conditions. As shown in the condition examples of Fig.~\ref{fig:system_overview_spacious}, we systematically randomize the time of day and weather effects for each scene.

\begin{figure*}[t]
  \centering
  \includegraphics[width=\linewidth]{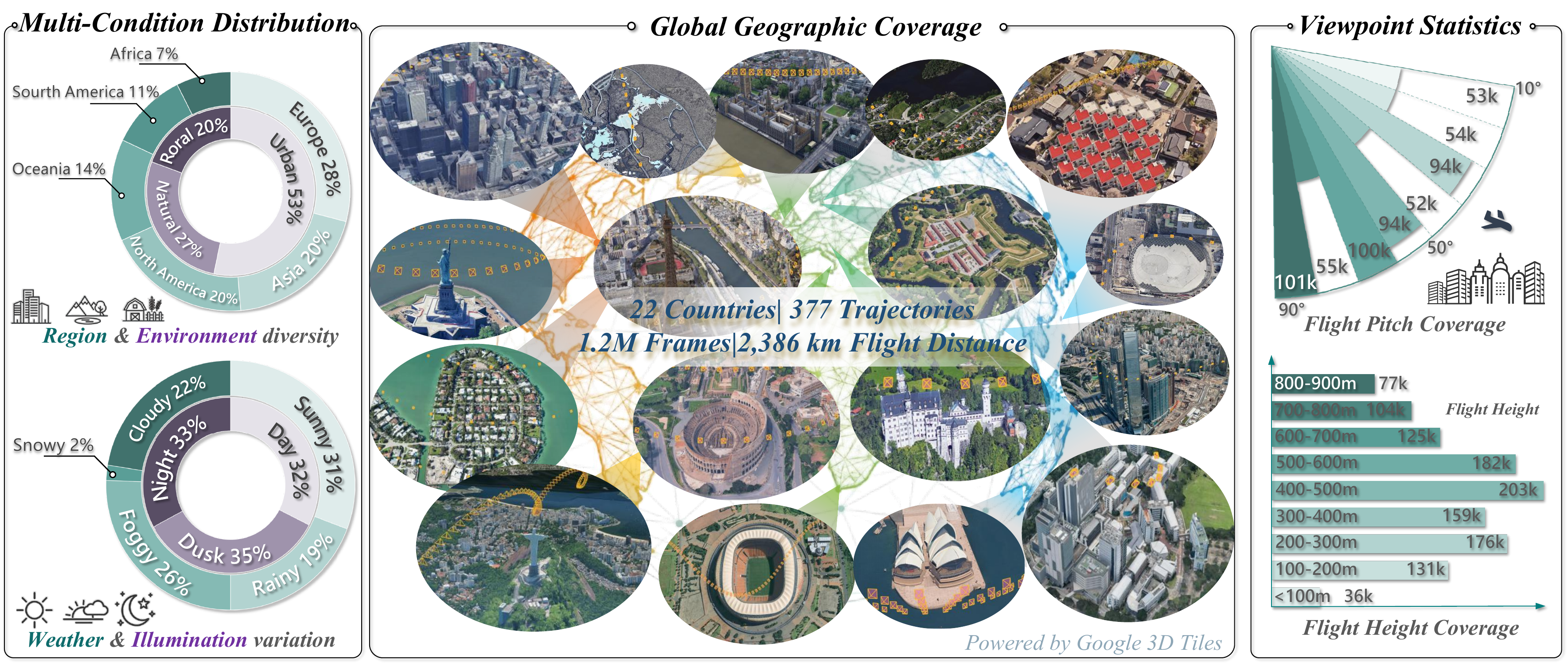}
  \caption{\textbf{Comprehensive statistics of AirZoo.} (Left) Region, environment, and weather distributions. (Center) Global coverage. (Right) Altitude and pitch envelopes.}
\label{fig:data_statistics}
\end{figure*}

\subsection{Dataset Statistics}

\noindent\textbf{Scale and Modalities.}
AirZoo represents the largest and most geographically diverse UAV geometric dataset to date, structured as continuous video sequences rather than isolated image collections. As shown in Fig.~\ref{fig:data_statistics}, it spans 377 trajectories across 22 countries, yielding over 1.2 million high-resolution ($1600 \times 1200$) frames derived from nearly 2,400\,km of cumulative flight trajectories. These regions cover six continents and encompass a wide range of environmental diversity, including urban, rural, and natural landscapes (summarized in the top-left of Fig.~\ref{fig:data_statistics}).

As illustrated in the left panel of Fig.~\ref{fig:data_statistics}, the sequences are annotated with semantic tags covering various weather presets (\textit{Sunny}, \textit{Cloudy}, \textit{Rainy}, \textit{Foggy}, \textit{Snowy}) and times of day (\textit{Day}, \textit{Sunset}, \textit{Night}). Along these sequences, our simulated acquisition flights operate within the 0--800\,m UAV altitude envelope. To provide oblique-to-nadir coverage for existing methods, we dynamically sweep the camera gimbal pitch from $10^{\circ}$ (oblique) to $90^{\circ}$ (nadir) across the sequences (detailed in the right panel of Fig.~\ref{fig:data_statistics}).

\noindent\textbf{Geometric Consistency.}
For seamless integration with standard geometric pipelines (e.g., COLMAP), we unify data structure and provide a standard pinhole camera model with per-frame intrinsics ($f_x, f_y, c_x, c_y$) and precise 6-DoF poses logged in both WGS84 (longitude, latitude, altitude) and ECEF Cartesian coordinate systems. The full pixel-to-WGS84 coordinate transformation chain is detailed in \cref{sec:supp-dataset,fig:projection_pipeline}.

\begin{figure}[t!]
\centering
\includegraphics[width=\linewidth]{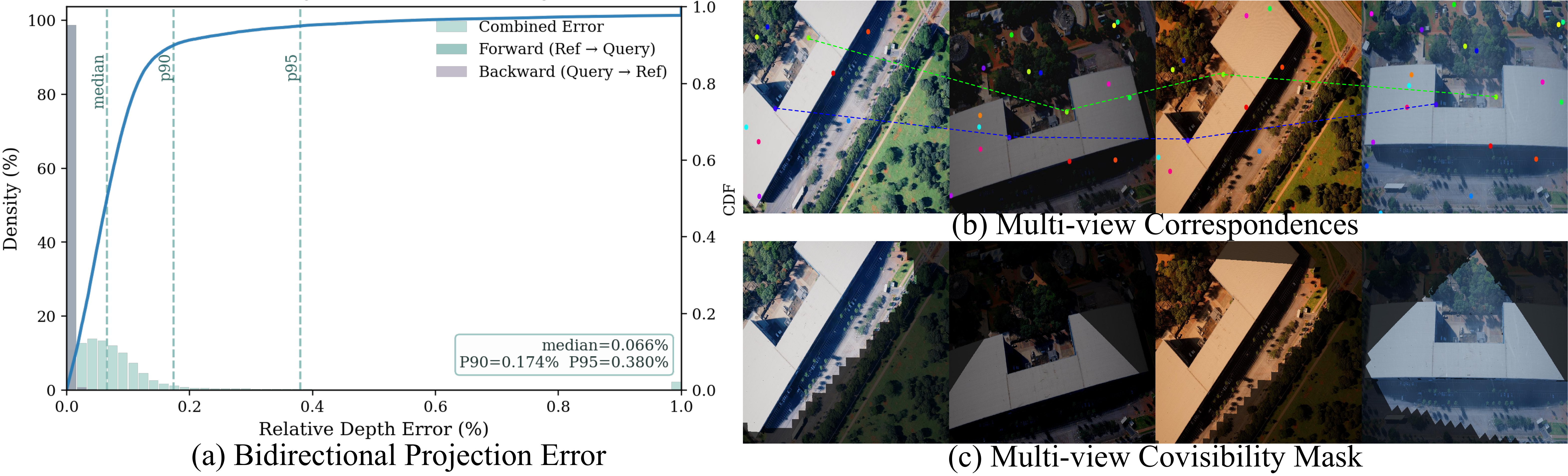}
\caption{\textbf{Geometric verification.} (a) Projection-error CDF with 0.066\% median error. (b,c) Multi-view correspondences and covisibility from one region across four weather trajectories.}
\label{fig:data_geometric_validation}
\end{figure}

To validate spatial fidelity, we back-project points from a source frame using rendered depth and pose, then re-project them into a neighboring target frame. Bidirectional projection yields a median relative depth error of 0.066\% with P90 at 0.174\% and P95 at 0.380\% (Fig.~\ref{fig:data_geometric_validation}), confirming pixel-level depth and pose fidelity.
We further visualize multi-view correspondences and covisibility across weather trajectories in Fig.~\ref{fig:data_geometric_validation}(b,c), showing stable geometric alignment under varying appearance conditions.

\begin{figure}[t]
\centering
\includegraphics[width=0.99\linewidth]{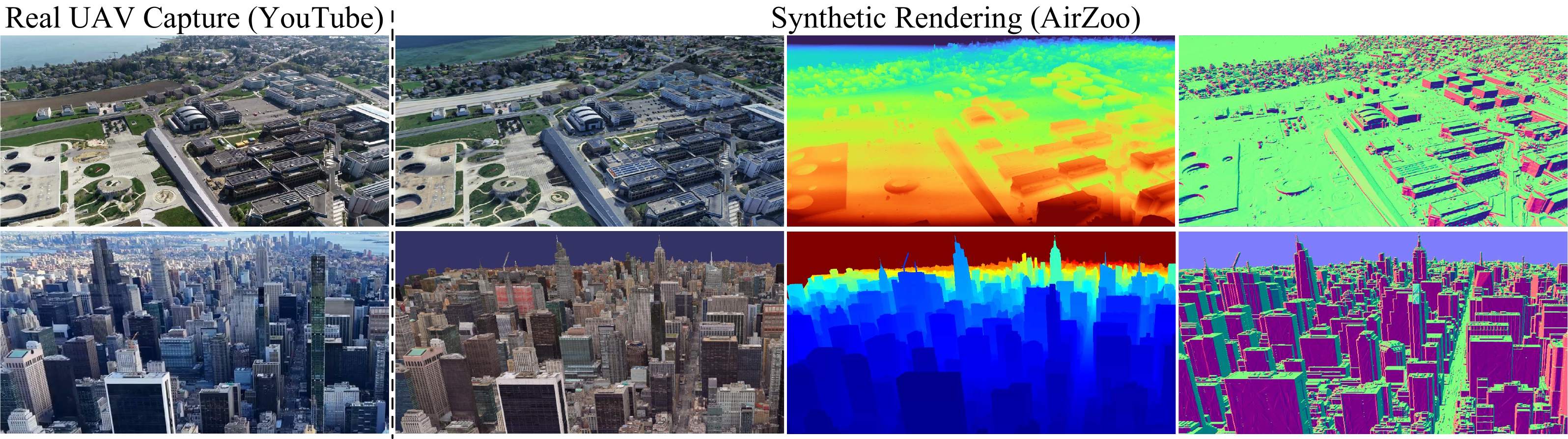}
\caption{\textbf{Real vs. synthetic comparison} at the same viewpoint.}
\label{fig:real_syn_comparison}
\end{figure}

\noindent\textbf{Photometric Realism.}
AirZoo uses Google Photorealistic 3D Tiles as the data source, rendering from real-world 3D meshes textured with high-resolution imagery~\cite{google_photorealistic_3d_tiles}. To inspect visual realism, we collect real UAV views from public DJI clips on YouTube, and reproduce the same viewpoints in AirZoo for synthetic rendering. As shown in Fig.~\ref{fig:real_syn_comparison}, the real and synthetic views are comparable in scene layout, object scale, and dominant structural patterns. Although lighting and shading differences remain visible cues for distinguishing the two, AirZoo additionally provides synchronized geometry that is difficult to obtain from real online videos.

\subsection{Dataset Usage and Evaluation Protocol}
We enforce zero spatial overlap between the AirZoo training set and all test benchmarks (Table~\ref{tab:dataset_usage}). Specifically, the synthetic AirZoo training split covers 19 countries, while Brazil, the USA, and New Zealand are held out for AirZoo-Test. The evaluation protocol combines AirZoo-Real with public UAV benchmarks, allowing us to test whether synthetic geometric diversity transfers to unseen real regions across retrieval, matching, and reconstruction.

\begin{table}[t]
    \centering
    \scriptsize
    \caption{Train/test datasets with zero spatial overlap between Part~1 and Part~2.}
    \label{tab:dataset_usage}
    \setlength{\tabcolsep}{3pt}
    \resizebox{\linewidth}{!}{%
    \begin{tabular}{l c l c c c}
    \toprule
    \textbf{Dataset} & \textbf{Type} & \textbf{Regions / Cities} & \textbf{Usage} & \textbf{Task} & \textbf{Results} \\
    \midrule
    \multicolumn{6}{@{}l}{\textbf{\textit{Part 1, Training (Synthetic)}}} \\
    AirZoo (Train) & Synth. & Global (19 countries) & Train & All & - \\
    \midrule
    \multicolumn{6}{@{}l}{\textbf{\textit{Part 2, Zero-Shot Test}}} \\
    AirZoo-Real      & Real & Changsha (China) & Test & All & Tabs.~\ref{tab:ourflight-retrieval}, \ref{tab:translation_accuracy}, \ref{tab:uav-recon} \\
    UAV-VisLoc~\cite{xu2024uav}       & Real & 11 regions (China) & Test & Retrieval & Tab.~\ref{tab:uav-visloc} \\
    AerialExtreMatch~\cite{aerialextrematch_localization_dataset} & Real & Changsha (China) & Test & Matching & Tab.~\ref{tab:6dof_recall} \\
    UAVScenes~\cite{wang2025uavscenes}        & Real & HK, Singapore & Test & Reconstruction & Tab.~\ref{tab:uav-recon} \\
    UrbanScene3D~\cite{lin2022capturing}     & Real & Shenzhen (China) & Test & Reconstruction & Tab.~\ref{tab:uav-recon} \\
    AirZoo-Test      & Synth. & Brazil, USA, NZ & Test & Reconstruction & Tab.~\ref{tab:uav-recon} \\
    \bottomrule
    \end{tabular}%
    }
\end{table}

\section{Experiments}
\label{sec:experiments}
We evaluate AirZoo as a training dataset on three representative UAV tasks: aerial image retrieval, cross-view matching, and multi-view 3D reconstruction. For each task, we keep the backbone architecture fixed and compare its original pre-trained version with the same model fine-tuned on AirZoo, so the performance gap reflects the data contribution rather than architectural changes.

\subsection{Aerial Image retrieval}
Aerial image retrieval is formulated as cross-view geo-localization: given a UAV query image, the goal is to retrieve the most relevant geo-tagged satellite tile. This setting is challenging because viewpoint, scale, illumination, and seasonal appearance can vary simultaneously. Existing methods still suffer from a clear domain gap. Several approaches~\cite{wu2024uavd4l, luo2024jointloc, he2024aerialvl} transfer models trained on ground-level data, and zero-shot pipelines such as AnyLoc~\cite{keetha2023anyloc} rely on frozen features. Game4Loc~\cite{ji2025game4loc} is designed for aerial retrieval but is trained in relatively limited synthetic environments.

\myparagraph{Training on AirZoo.}
We adopt MegaLoc~\cite{berton2025megaloc} as the base retriever and fine-tune it on AirZoo. MegaLoc is pre-trained on four ground-view datasets~\cite{ali2022gsv,warburg2020mapillary,tung2024megascenes,dai2017scannet}, providing strong generic representations but remains suboptimal for nadir-to-oblique matching. During fine-tuning, we use weighted InfoNCE~\cite{ji2025game4loc}, where positive similarity is modulated by continuous overlap ratios between queries and satellite tiles. 

\myparagraph{Evaluation Benchmark.}
We evaluate on two real-world aerial benchmarks: the public UAV-VisLoc~\cite{xu2024uav} and AirZoo-Real (i.e., our newly collected real-flight set). In both settings, UAV images serve as queries, and the retrieval gallery is built from Google Maps~\cite{googlemaps} satellite imagery at 0.3\,m ground resolution. Together, the benchmark covers 13 geographic regions and over 1,000 query images across diverse lighting conditions and flight altitudes. To reduce scale ambiguity caused by altitude variation, we construct the gallery with multi-scale cropping (tile sizes $512\times512$ and $1024\times1024$) and a 50\% overlap ratio. 

\myparagraph{Evaluation Metrics.}
We use Recall@K ($K\in\{1,3,5\}$) as the primary metric, where a query is counted as correct if at least one positive tile appears in the top-$K$ retrieved results. We also report AP@5 to evaluate ranking quality among the top returned candidates.

\begin{table}[t]
\centering
\scriptsize
\caption{\textbf{Geo-localization results on the UAV-VisLoc dataset. }
Our AirZoo fine-tuned MegaLoc consistently improves over the original MegaLoc across all evaluation metrics (R@1, R@3, R@5, and AP@5). 
Improvements over MegaLoc are shown in gray, and the best results are highlighted in \textbf{bold}.}
\setlength{\tabcolsep}{4pt} % reduce column spacing
\renewcommand{\arraystretch}{1.1} % slightly tighter rows
\begin{tabular}{lcccc}
\toprule
\textbf{Method} & \textbf{R@1} & \textbf{R@3} & \textbf{R@5} & \textbf{AP@5} \\
\midrule
AnyLoc~\cite{keetha2023anyloc} & 17.01 & 29.51 & 36.57 & 12.60 \\
SALAD~\cite{izquierdo2024optimal} & 25.72 & 40.03 & 46.77 & 17.95 \\
Game4Loc~\cite{ji2025game4loc} & 18.68 & 30.75 & 41.79 & 13.20 \\
MegaLoc~\cite{berton2025megaloc} & 26.91 & 45.04 & 52.68 & 19.09 \\
\rowcolor{yellow!25}
MegaLoc (Ours) & 
\textbf{28.64} {\color{gray}(+\textbf{1.73})} & 
\textbf{46.33} {\color{gray}(+\textbf{1.29})} & 
\textbf{53.74} {\color{gray}(+\textbf{1.06})} & 
\textbf{19.78} {\color{gray}(+\textbf{0.69})} \\
\bottomrule
\end{tabular}
\label{tab:uav-visloc}
\end{table}

\begin{table}[t]
\centering
\scriptsize
\caption{\textbf{Geo-localization results on the AirZoo-Real dataset.} 
Our AirZoo fine-tuned MegaLoc consistently improves over the original MegaLoc across all evaluation metrics (R@1, R@3, R@5, and AP@5). 
Improvements over MegaLoc are shown in gray, and the best results are highlighted in \textbf{bold}.}
\setlength{\tabcolsep}{4pt}
\renewcommand{\arraystretch}{1.1}
\begin{tabular}{lcccc}
\toprule
\textbf{Method} & \textbf{R@1} & \textbf{R@3} & \textbf{R@5} & \textbf{AP@5} \\
\midrule
AnyLoc~\cite{keetha2023anyloc} & 4.52 & 9.35 & 10.45 & 2.94 \\
SALAD~\cite{izquierdo2024optimal} & 7.36 & 19.18 & 29.48 & 7.73 \\
Game4Loc~\cite{ji2025game4loc} & 12.67 & 23.59 & 34.18 & 10.66 \\
MegaLoc~\cite{berton2025megaloc} & 6.46 & 16.52 & 22.16 & 6.33 \\
\rowcolor{yellow!25}
MegaLoc(Ours) & 
\textbf{18.66} {\color{gray}(+\textbf{12.21})} &
\textbf{40.70} {\color{gray}(+\textbf{24.18})} &
\textbf{50.53} {\color{gray}(+\textbf{28.38})} &
\textbf{17.22} {\color{gray}(+\textbf{10.89})} \\
\bottomrule
\end{tabular}
\label{tab:ourflight-retrieval}
\end{table}

\begin{figure*}[t]
  \centering
  \includegraphics[width=\linewidth]{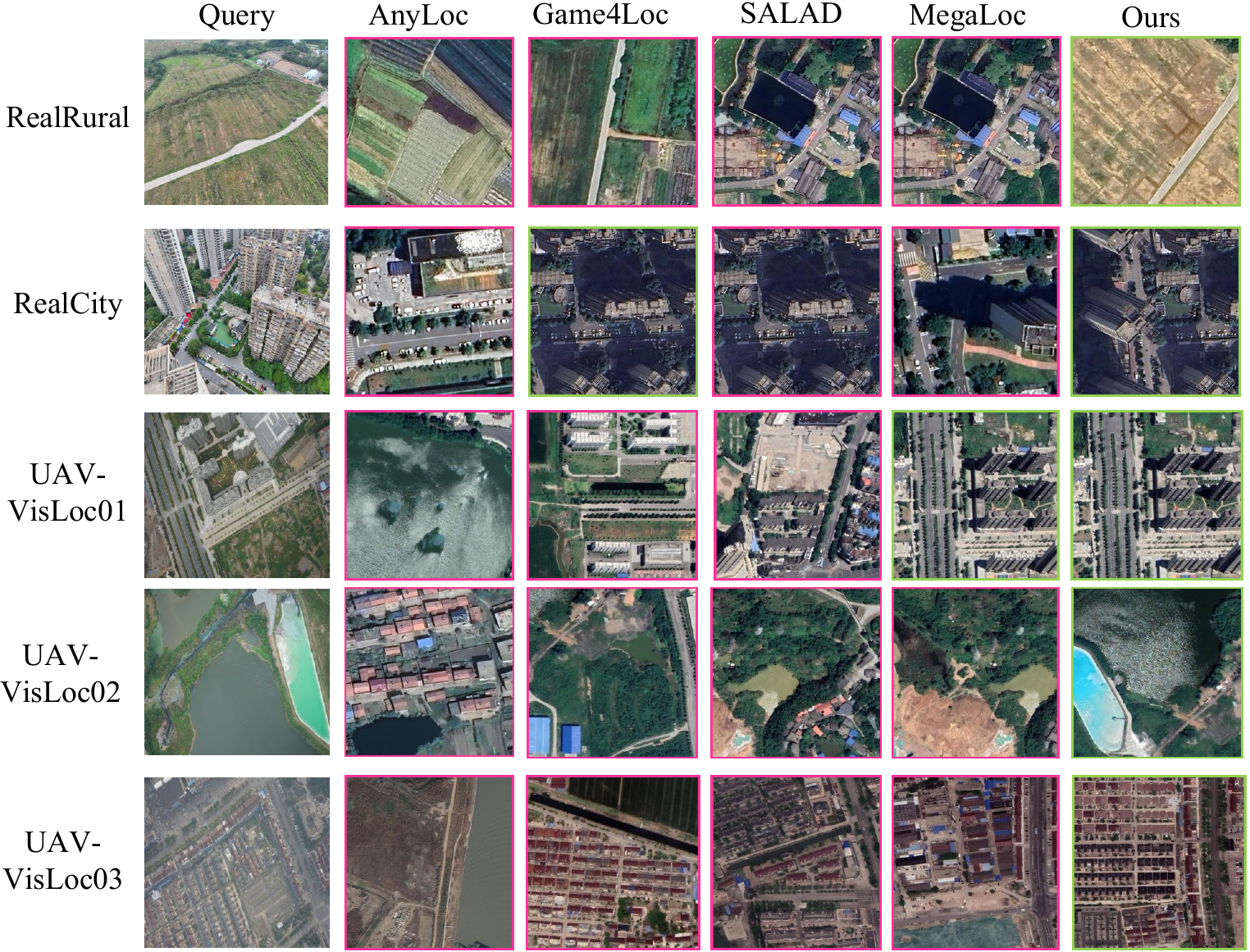}
  \caption{\textbf{Qualitative cross-view geo-localization comparisons.}
  Each row corresponds to a real-world UAV query, and each column shows the Top-1 retrieved satellite tile from different methods.
  Correct and incorrect predictions are highlighted with green and pink borders, respectively.
  Fine-tuning on AirZoo yields more stable retrievals across challenging viewpoints and environmental changes.}
  \label{fig:retrieval_results}
\end{figure*}

\myparagraph{Results.} Quantitative comparisons on UAV-VisLoc and our AirZoo-Real are reported in Tab.~\ref{tab:uav-visloc} and Tab.~\ref{tab:ourflight-retrieval}, respectively (with per-scene results in \cref{tab:uav_visloc_scenes,tab:uav_real_scenes}). We observe that 1) AirZoo fine-tuning consistently improves top-rank retrieval quality across both benchmarks; 2) the gain on UAV-VisLoc is moderate since its query-reference pairs are predominantly nadir-dominant and primarily differ in scale rather than viewpoint, making the cross-view gap less pronounced; and 3) the gain on our real-flight benchmark is substantially larger, where arbitrary viewpoints coupled with stronger environmental variations create a more challenging cross-view domain gap that better reflects real-world deployment scenarios. These results confirm that AirZoo is particularly effective for improving robustness in realistic UAV geo-localization under extreme cross-view conditions beyond near-nadir settings.

\subsection{Cross-view Matching}
Cross-view matching between orthophoto references and oblique UAV imagery is a core step for UAV pose estimation. The task is difficult because large viewpoint changes, altitude variation, and weather differences jointly affect geometric correspondence quality. Most existing matchers are not designed for extreme aerial gaps. ELoFTR~\cite{wang2024efficient} and RoMa~\cite{edstedt2024roma} are primarily trained on ground-view datasets such as MegaDepth~\cite{li2018megadepth}. RoMa (GIM)~\cite{xuelun2024gim} leverages massive video supervision to enlarge data scale; however, such video-derived data lacks long-term temporal spans and diverse environmental dynamics, failing to capture complex illumination.

\myparagraph{Training on AirZoo.}
We adopt RoMa~\cite{edstedt2024roma} as the base matcher and fine-tune it on a curated mixture of the MegaDepth~\cite{li2018megadepth} and AirZoo datasets. The network weights are initialized from the publicly available RoMa checkpoint pre-trained on outdoor scenes~\cite{edstedt2024roma}.

\myparagraph{Evaluation Benchmark.}
We evaluate on two complementary benchmarks: the public AerialExtreMatch~\cite{aerialextrematch_localization_dataset} dataset and AirZoo-Real (i.e., our newly collected real-flight benchmark). Combined, they span three geographic locations and more than 500 UAV query images. For each query, we use its GPS prior to crop spatially corresponding reference tiles from Digital Surface/Orthophoto Models (DSM/DOM), and then perform cross-view matching on these candidates.

\myparagraph{Evaluation Metrics.}
We report recall-based localization metrics. On AerialExtreMatch, which provides full 6-DoF annotations, we measure pose recall at $(5\text{m}, 1^\circ)$, $(10\text{m}, 1^\circ)$, and $(20\text{m}, 2^\circ)$. On the AirZoo-Real dataset, where the ground truth comes from precise RTK positions, we report translation recall at $5\text{m}$, $10\text{m}$, and $20\text{m}$, together with median translation error.

\begin{table}[t]
\centering
\scriptsize
\caption{\textbf{6-DoF localization results on the AerialExtreMatch dataset.} 
Gray values indicate improvements over the previous baseline.}
\setlength{\tabcolsep}{5pt}
\renewcommand{\arraystretch}{1.1}
\begin{tabular}{lccc}
\toprule
\textbf{Method} & \textbf{(5m, 1°) $\uparrow$} & \textbf{(10m, 1°) $\uparrow$} & \textbf{(20m, 2°) $\uparrow$} \\
\midrule
LoFTR~\cite{sun2021loftr} & 66.67 & 66.67 & 84.47 \\
ELoFTR~\cite{wang2024efficient} & 81.82 & 81.82 & 85.61 \\
DUSt3R~\cite{wang2024dust3r} & 1.52 & 3.78 & 16.29 \\
MASt3R~\cite{murai2025mast3r} & 76.14 & 76.52 & 87.50 \\
RoMA~\cite{edstedt2024roma} & 95.83 & 95.83 & 96.59 \\
\rowcolor{brown!15}
RoMA (GIM)~\cite{xuelun2024gim} & 
\textbf{94.32} {\color{gray}(--\textbf{1.51})} & 
\textbf{94.32} {\color{gray}(--\textbf{1.51})} & 
\textbf{97.35} {\color{gray}(+\textbf{0.76})} \\
\rowcolor{yellow!25}
RoMA (Ours) & 
\textbf{96.21} {\color{gray}(+\textbf{0.38})} & 
\textbf{96.21} {\color{gray}(+\textbf{0.38})} & 
\textbf{98.11} {\color{gray}(+\textbf{1.52})} \\
\bottomrule
\end{tabular}
\label{tab:6dof_recall}
\end{table}

\begin{table}[t]
\centering
\scriptsize
\caption{\textbf{Translation results on the AirZoo-Real dataset}. Gray values indicate improvements over the previous baseline.}
\setlength{\tabcolsep}{5pt}
\renewcommand{\arraystretch}{1.1}
\begin{tabular}{lcccc}
\toprule
\textbf{Method} & \textbf{median} $\downarrow$ & \textbf{acc@5m} $\uparrow$ & \textbf{acc@10m} $\uparrow$& \textbf{acc@20m} $\uparrow$ \\
\midrule
LoFTR~\cite{sun2021loftr} & 244.20 & 3.04 & 6.89 & 8.13 \\
ELoFTR~\cite{wang2024efficient} & 229.64 & 3.93 & 7.73 & 9.40 \\
DUSt3R~\cite{wang2024dust3r} & 214.57 & 2.46 & 7.80 & 14.15 \\
MASt3R~\cite{murai2025mast3r} & 183.75 & 12.70 & 20.06 & 24.12 \\
RoMA~\cite{edstedt2024roma} & 5.87 & 44.95 & 55.58 & 57.30 \\
\rowcolor{brown!15}
RoMA (GIM)~\cite{xuelun2024gim} & 
\textbf{3.22} {\color{gray}(--\textbf{2.65})} &
\textbf{77.71} {\color{gray}(+\textbf{32.76})} &
\textbf{84.14} {\color{gray}(+\textbf{28.56})} &
\textbf{86.93} {\color{gray}(+\textbf{29.63})} \\
\rowcolor{yellow!25}
RoMA (Ours) & 
\textbf{3.03} {\color{gray}(--\textbf{2.84})} &
\textbf{79.78} {\color{gray}(+\textbf{34.83})} &
\textbf{85.35} {\color{gray}(+\textbf{29.77})} &
\textbf{86.93} {\color{gray}(+\textbf{29.63})} \\
\bottomrule
\end{tabular}
\label{tab:translation_accuracy}
\end{table}

\begin{figure*}[t]
  \centering
  \includegraphics[width=\linewidth]{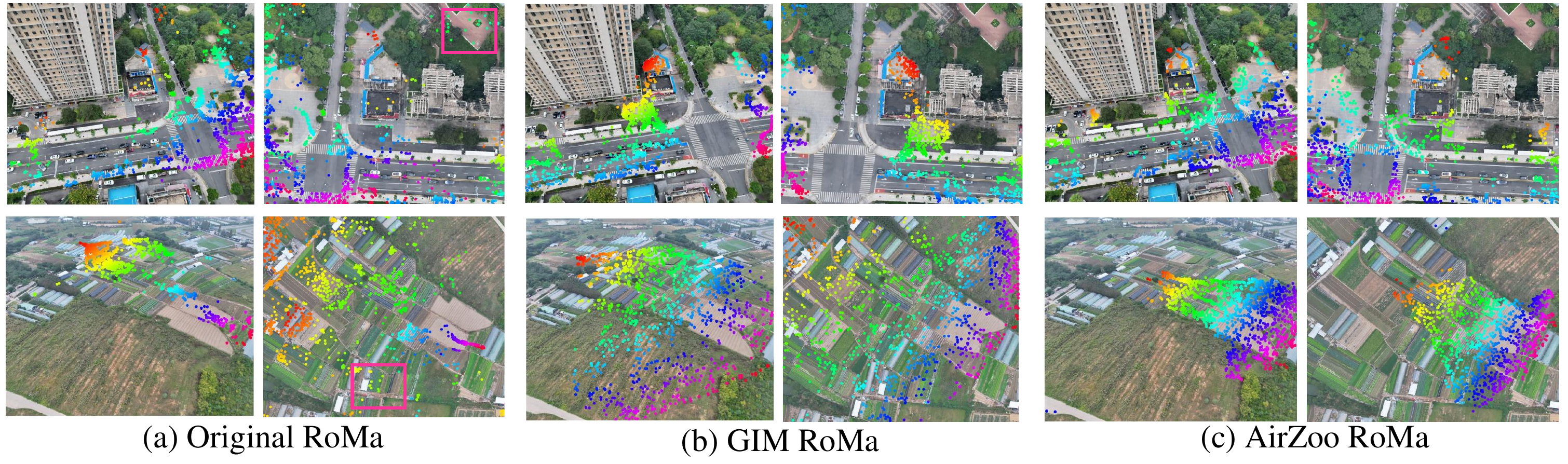}
  \caption{\textbf{Qualitative cross-view matching results on AirZoo-Real.} Pink boxes highlight mismatched regions. The original RoMa produces many incorrect correspondences under aerial cross-view conditions. RoMa (GIM), trained on large-scale aerial video data, alleviates this issue to some extent. Our RoMa, trained specifically for cross-view matching, produces more uniformly distributed correspondences concentrated on static scene structures.}
  \label{fig:match_results}
\end{figure*}

\myparagraph{Results.} Quantitative evaluations on the AerialExtreMatch and AirZoo-Real datasets are reported in Tab.~\ref{tab:6dof_recall} and Tab.~\ref{tab:translation_accuracy}, with qualitative comparisons in Fig.~\ref{fig:match_results}. We observe that 1) AirZoo fine-tuning consistently improves performance over the original RoMa in 6-DoF and translation-only evaluations; 2) the advantage remains under both strict and relaxed error thresholds, indicating stable geometric generalization rather than threshold-specific gains; and 3) improvements are especially clear in translation-only localization, where the AirZoo-trained model produces fewer mismatches and more spatially distributed correspondences on static scene regions. AirZoo provides effective cross-view supervision for robust UAV matching and downstream pose estimation under challenging viewpoint and environmental variations.

\begin{figure*}[!t]
  \centering
  \includegraphics[width=\linewidth]{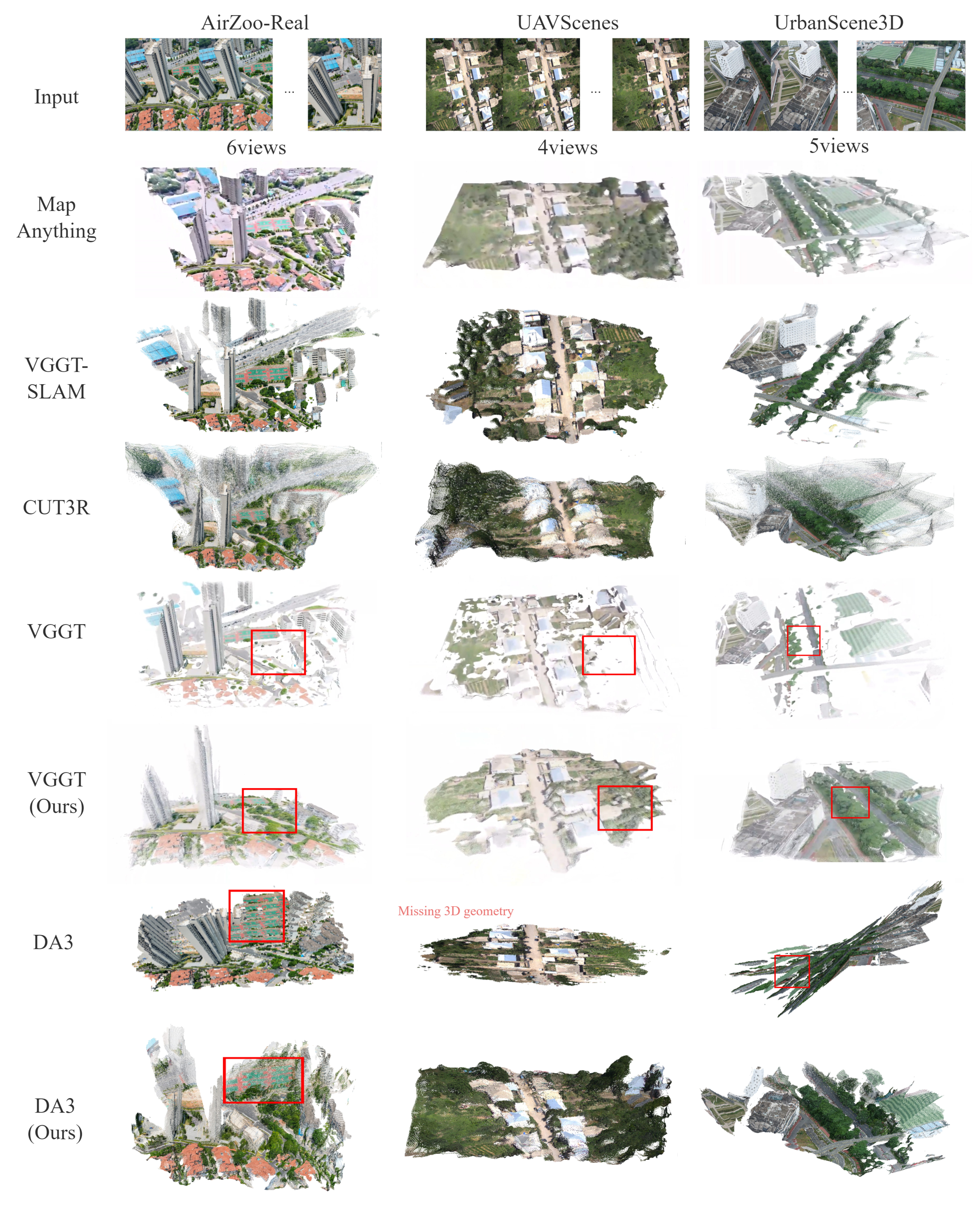}
  \caption{\textbf{Qualitative results on AirZoo-Real, UAVScenes and UrbanScene3D.} Red boxes highlight missing or erroneous regions in the reconstruction produced by the original model. For each column, the reconstruction result is shown with a side view on the left and a top-down view on the right. Compared with the baselines, our fine-tuned model produces cleaner and more complete geometry.}
  \label{fig:recon_results}
\end{figure*}

\subsection{Multi-view 3D Reconstruction}
Multi-view 3D reconstruction from UAV imagery aims to recover camera poses and dense geometry from image sequences. Compared with ground-view capture, UAV data introduces larger altitude changes, stronger oblique views, and wider illumination variations.
Recent transformer-based systems, including CUT3R~\cite{wang2025continuous}, MapAnything~\cite{keetha2025mapanything}, VGGT~\cite{wang2025vggt}, and DA3~\cite{lin2025depth}, are mainly trained on ground-level datasets such as CO3D~\cite{reizenstein21co3d} and RealEstate10K~\cite{46965}. As a result, their direct transfer to UAV trajectories remains challenging.

\myparagraph{Training on AirZoo.}
We fine-tune VGGT~\cite{wang2025vggt} and DA3~\cite{lin2025depth} on AirZoo. We select 16 trajectories from Brazil, the USA, and New Zealand for validation/testing, and use the remaining 361 trajectories for training on 8 A100 GPUs. During training, we sample clips with overlap ratios between 50\% and 75\%, and variable sequence lengths (2--24 frames for VGGT, 2--10 frames for DA3). This strategy increases exposure to diverse baselines and motion patterns.

\myparagraph{Evaluation Benchmark.}
We evaluate on a synthetic test set constructed by selecting 16 trajectories from AirZoo, as well as on AirZoo-Real (i.e., our newly collected real-flight set). The real-flight dataset contains 9,430 images captured over four areas (school, driving school, substation, and plaza) during three time windows (06:00--08:00, 12:00--14:00, and 18:00--20:00), with additional low-light data (22:00--24:00) for two scenes. During acquisition, the UAV pitch angle is maintained between 30° and 45° at approximately 160 m altitude, with RTK ensuring accurate extrinsics. Depth maps are obtained by rendering from 3D models built via oblique photogrammetry.

\myparagraph{Evaluation Metrics.}
Following the standard evaluation protocol in~\cite{lin2025depth}, we report distance-thresholded 3D reconstruction metrics using the F1 score computed from precision and recall based on Chamfer Distance, with a threshold of $d=10$.

\begin{table}[t]
\centering
\scriptsize
\caption{\textbf{Reconstruction results on AirZoo-Test, AirZoo-Real, UAVScenes and UrbanScene3D datasets. }
Our AirZoo fine-tuned VGGT and DA3 consistently improve over the original methods across all testsets. 
Improvements over original methods are shown in gray, and the best results are highlighted.}
\setlength{\tabcolsep}{4pt} % reduce column spacing
\renewcommand{\arraystretch}{1.1} % slightly tighter rows
\begin{tabular}{lcccc}
\toprule
\multirow{2}{*}{\textbf{Method}}
& \textbf{UAVScenes~\cite{wang2025uavscenes}}
& \textbf{UrbanScene3D~\cite{lin2022capturing}}
& \textbf{AirZoo-Test}
& \textbf{AirZoo-Real} \\
& \textbf{F1$\uparrow$} & \textbf{F1$\uparrow$} & \textbf{F1$\uparrow$} & \textbf{F1$\uparrow$} \\
\midrule
CUT3R~\cite{wang2025continuous} &83.51 &39.64& 64.99 & 48.88 \\
MapAnything~\cite{keetha2025mapanything} &90.30& 49.73 & 83.35 & 45.00 \\
VGGT-SLAM~\cite{maggio2025vggt} &86.42 &45.32& 75.65 & \textbf{53.95} \\
VGGT~\cite{wang2025vggt} &86.20&43.38& 67.67 & 42.48 \\
VGGT(Ours) &86.98{\color{gray}(+\textbf{0.78})}&52.61{\color{gray}(+\textbf{8.23})}& 77.32{\color{gray}(+\textbf{9.65})} & 43.67{\color{gray}(+\textbf{1.19})} \\
DA3-Large~\cite{lin2025depth} &87.38& 47.83 & 54.10 & 49.97 \\
\rowcolor{yellow!25}
DA3-Large (Ours) & 
\textbf{91.42} {\color{gray}(+\textbf{2.93})} & 
\textbf{55.91} {\color{gray}(+\textbf{ 8.08})} & 
\textbf{86.09} {\color{gray}(+\textbf{31.99})} & 
53.07 {\color{gray}(+\textbf{3.10})} \\
\bottomrule
\end{tabular}
\label{tab:uav-recon}
\end{table}

\myparagraph{Results.}
Quantitative evaluations on UAVScenes~\cite{wang2025uavscenes}, UrbanScene3D~\cite{lin2022capturing}, AirZoo test set, and AirZoo-Real are reported in Tab.~\ref{tab:uav-recon}, with qualitative comparisons in Fig.~\ref{fig:recon_results}. We observe that 1) AirZoo fine-tuning consistently improves performance over both VGGT and DA3 across all test sets; 2) the discrepancy between real-world and synthetic results reveals that domain gap remains a substantial challenge for real-flight trajectories, where AirZoo training provides notable improvements. We also note that while VGGT-SLAM achieves the best performance on AirZoo-Real, this is largely due to its BA-based optimization, giving it an accuracy edge over purely feed-forward baselines. AirZoo provides effective supervision for robust multi-view 3D reconstruction from UAV sequences under challenging viewpoint, illumination, and environmental variations.

\section{Conclusion}
This paper presents AirZoo, a unified large-scale dataset and benchmark for grounding aerial geometric 3D vision. The proposed benchmark introduces three key contributions namely 1) a scalable generation pipeline that leverages world-scale photogrammetric meshes to render customizable UAV trajectories, 2) comprehensive scene diversity spanning 377 weather-conditioned trajectories across 22 countries, and 3) rich geometric annotations providing synchronized pixel-level metric depth and precise 6-DoF geo-referenced poses. Extensive evaluations across three rigorous geometric tracks demonstrate that AirZoo serves as a powerful pre-training engine, yielding substantial performance gains for state-of-the-art models on real-world benchmarks. We believe this work not only advances the field of aerial geometric perception by bridging the critical gap in high-fidelity training data, but also establishes a new performance upper bound for aerial spatial intelligence.

\section*{Acknowledgments}
{\small
This work was supported by the National Natural Science Foundation of China (NSFC) Young Scientists Fund (Grant Nos.~62406331 and 62503491).
The authors would like to thank Yipeng Liu, Yuxuan Cheng, Wanchang Li , Qing Chang for dataset construction, Xia Li for rendering engine debugging.
We sincerely thank \textcolor{eccvblue}{\href{https://cesium.com/platform/cesium-for-unreal/}{Cesium for Unreal}}, \textcolor{eccvblue}{\href{https://microsoft.github.io/AirSim/}{AirSim}}, and \textcolor{eccvblue}{\href{https://earth.google.com/web/}{Google Earth}} for providing the data platform, simulation framework, and data source.
}

\bibliographystyle{splncs04}
\bibliography{main}

\maketitlesupplementary
\section{Dataset Details}
\label{sec:supp-dataset}

\subsection{Simulator Implementation}
An overview of the AirSim-Cesium-Unreal simulator is provided in \cref{sec:dataset,fig:pipeline_workflow}. Below we detail the external and internal control environments. The system operates through two primary environments. The external control environment (Python) acts as the master controller. It begins by parsing predefined flight trajectories (stored as JSON/CSV files containing WGS84 coordinates and specific Euler angles). Through the AirSim Client RPC interface, it establishes a connection with the simulation engine, enabling API control to sequentially execute the waypoint queue. The main loop iteratively moves the virtual UAV, triggers synchronous sensor captures (RGB and depth), and logs the corresponding absolute poses.

The internal simulation and rendering environment (Unreal Engine) is responsible for the physical simulation and high-fidelity rendering. The core components include the \texttt{CesiumGeoreference} actor, which anchors the Unreal coordinate system to a specific geographic origin, and the \texttt{CesiumSunSky} controller, which governs the atmospheric and weather variations. The AirSim plugin physically instantiates the UAV vehicle and virtual cameras, converting the designated Field of View (FOV) into intrinsic matrices ($K$). To guarantee data alignment, the simulator operates in synchronous mode, pausing the physics tick until both the RGB and depth buffers are fully rendered and transmitted back to the Python client.

\subsection{Flight Trajectories Design and Dataset Characteristics}
\label{sup:flight_trajectories}

\begin{figure}[t!]
  \centering
  \includegraphics[width=\linewidth]{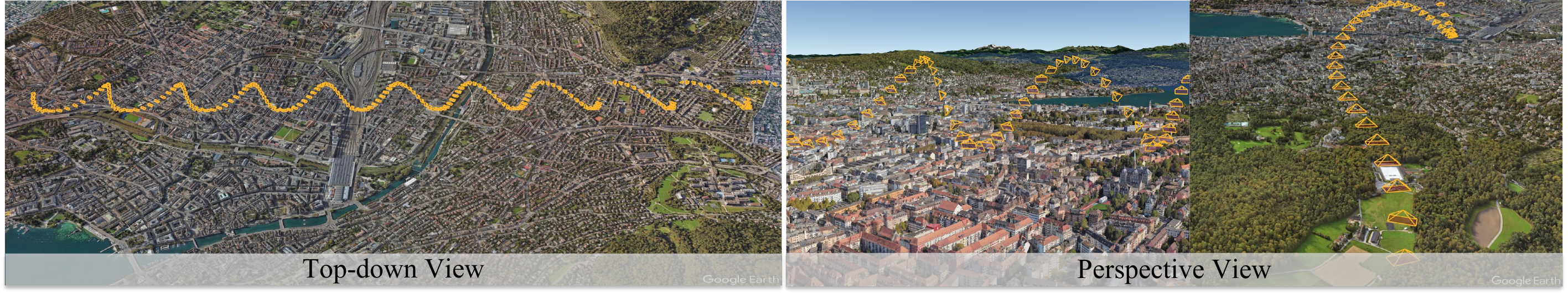}
  \caption{\textbf{Visualization of the designed ``barrel-roll'' trajectory.} The top-down view (top) shows yaw variation along the S-shaped path. The perspective views (bottom) illustrate changes in altitude (up-and-down arcs) and pitch (tilting of camera frustums).}
  \label{fig:supp_flight_design}
\end{figure}

\begin{figure*}[t!]
  \centering
  \includegraphics[width=\linewidth]{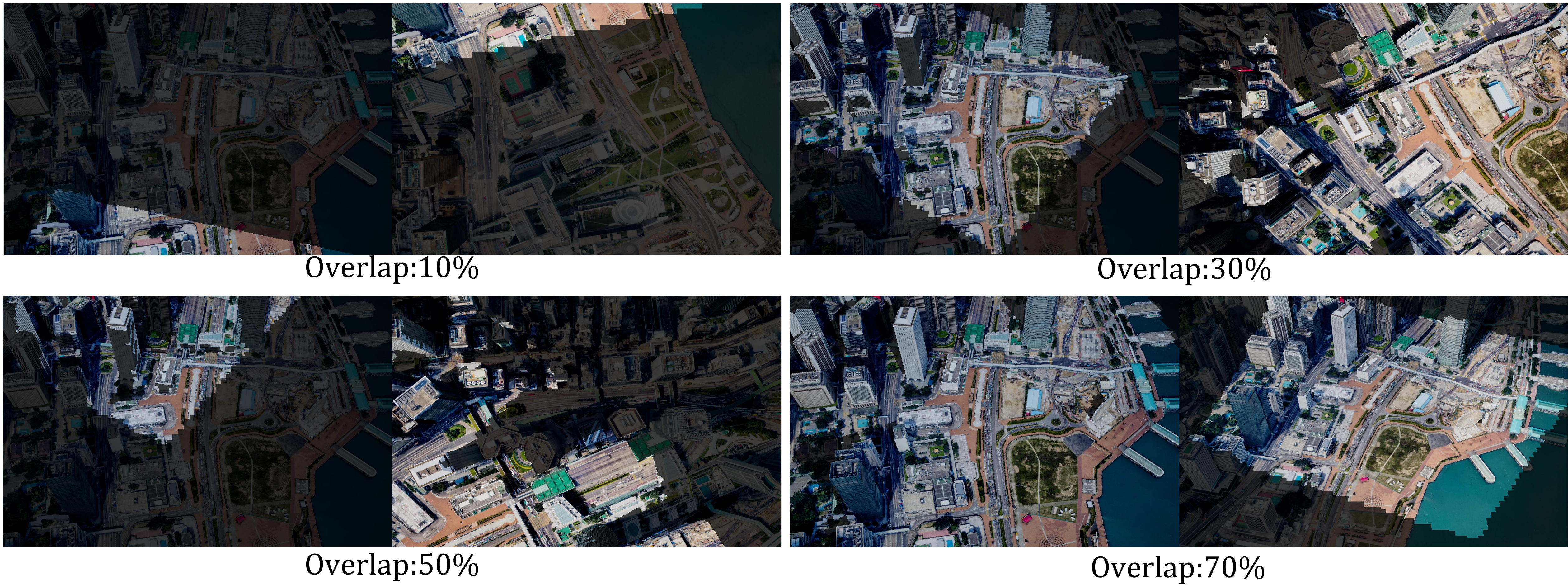}
  \caption{\textbf{Visualization of varying overlap ratios in AirZoo.} By carefully designing the UAV trajectories, we provide challenging image pairs with low visual overlap (down to 10\%), serving as a rigorous testbed for wide-baseline matching and localization tasks.}
  \label{fig:overlap_visualization}
\end{figure*}

\noindent\textbf{Flight Trajectories Generation.}
We design ``barrel-roll--inspired'' UAV trajectories that combine horizontal orbiting around a local scene center with straight cruising and curved sweeps, while altitude, yaw, and gimbal pitch are adjusted along the path (\cref{fig:supp_flight_design}). This yields both smooth sequential motion and challenging wide-baseline transitions, producing dense temporal neighborhoods for reconstruction alongside broader viewpoint changes for retrieval and matching. As visualized in \cref{fig:overlap_visualization}, the resulting image pairs span overlap ratios from dense sequential frames (70\%) to extreme wide-baseline scenarios (10\%).

\begin{figure*}[t!]
  \centering
  \includegraphics[width=\linewidth]{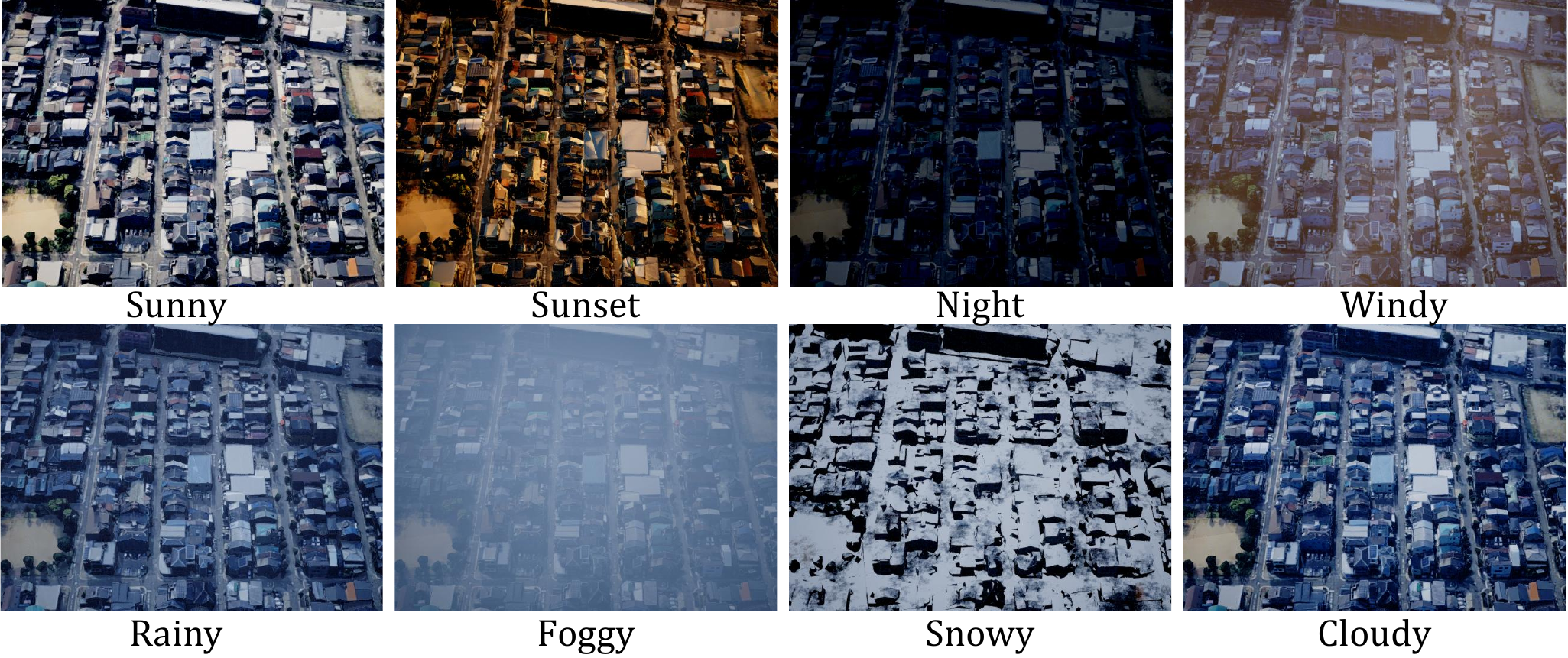}
  \caption{\textbf{Environmental diversity in AirZoo.} Examples of weather (\textit{Sunny}, \textit{Cloudy}, \textit{Rainy}, \textit{Foggy}, \textit{Snowy}) and illumination (\textit{Day}, \textit{Sunset}, \textit{Night}) variants rendered along the same trajectory.}
  \label{fig:weather_variants}
\end{figure*}

\noindent\textbf{Environmental Simulations.}
\cref{fig:weather_variants} shows representative renders along identical flight paths under the weather and illumination presets defined in the main paper: \textit{Sunny}, \textit{Cloudy}, \textit{Rainy}, \textit{Foggy}, \textit{Snowy}, and \textit{Day}, \textit{Sunset}, \textit{Night}. Fixing geometry while varying appearance enables evaluation under challenging lighting and weather conditions.

\begin{figure*}[t!]
  \centering
  \includegraphics[width=\linewidth]{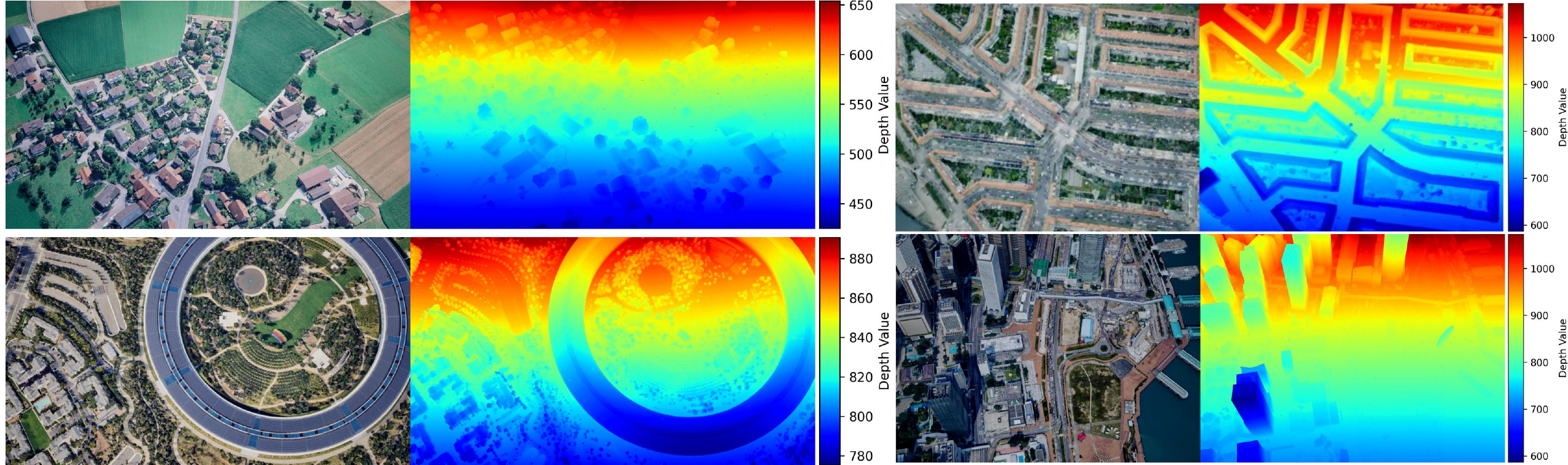}
  \caption{\textbf{RGB--depth pairs.} Each RGB image is accompanied by a pixel-aligned metric depth map. Colors indicate absolute distance.}
  \label{fig:depth_pairs}
\end{figure*}

\noindent\textbf{RGB-Depth Alignment.}
\cref{fig:depth_pairs} visualizes pixel-aligned RGB--depth pairs across urban and rural landscapes. Sharp depth boundaries around buildings and foliage illustrate the annotation quality; quantitative projection errors are reported in \cref{fig:data_geometric_validation}.

\subsection{Geographic Projection and Coordinate Transformation}

\begin{figure*}[t!]
  \centering
  \includegraphics[width=\linewidth]{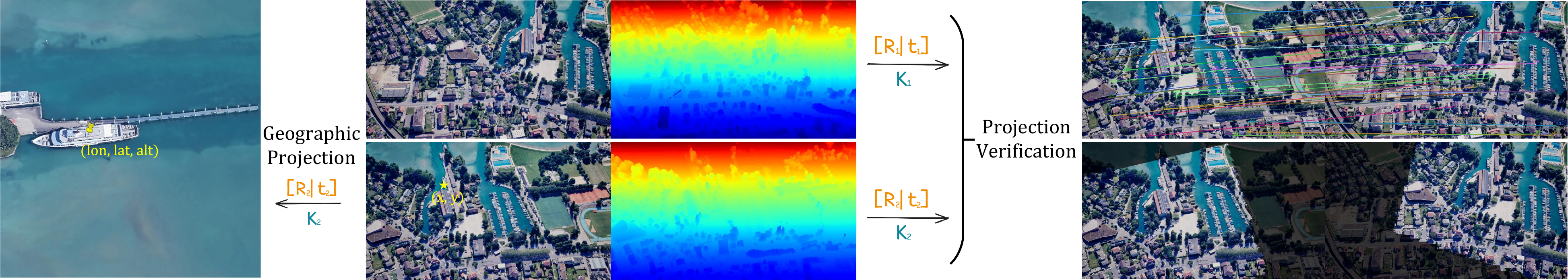}
  \caption{\textbf{Geographic projection and verification.} The left part illustrates the projection from 2D pixels to global WGS84 coordinates ($\text{lon, lat, alt}$). The middle part shows the rendered RGB--depth pairs. The right part demonstrates projection verification through epipolar geometry (top-right) and visual overlap (bottom-right).}
  \label{fig:projection_pipeline}
\end{figure*}

A critical challenge in generating synthetic geographic data is maintaining strict alignment between the simulator's local coordinate system and global geographic coordinates. AirZoo resolves this through a rigorous spatial transformation chain managed by the \texttt{CesiumGeoreference} service, as illustrated in \cref{fig:projection_pipeline}.

For any given pixel $\mathbf{p} = [u, v, 1]^T$ in the captured image, its corresponding 3D ray in the camera coordinate system is obtained via the inverse intrinsic matrix $K^{-1}$. The complete spatial chain from a 2D pixel to a global WGS84 coordinate is formulated as follows:
$$ \mathbf{P}_{cam} = K^{-1} \mathbf{p} \cdot z $$
where $z$ is the metric depth value. The point is then transformed into the local East-North-Up (ENU) coordinate system, and subsequently into the global Earth-Centered, Earth-Fixed (ECEF) system using the camera extrinsic matrices:
$$ \mathbf{P}_{ECEF} = \mathbf{T}_{ENU \rightarrow ECEF} \left( \mathbf{T}_{cam \rightarrow ENU} \mathbf{P}_{cam} \right) $$
Finally, the ECEF coordinates are projected into the standard WGS84 format (Longitude, Latitude, Altitude) via the standard ECEF-to-LLA conversion.

This rigorous chain ($\text{pixel} \xrightarrow{K^{-1}} \text{ray} \xrightarrow{T} \text{ENU} \xrightarrow{T} \text{ECEF} \xrightarrow{} \text{WGS84}$) ensures that all rendered depth maps and 6-DoF poses in AirZoo are geometrically aligned. To verify this alignment, we perform a projection verification process (see the rightmost part of \cref{fig:projection_pipeline}). By back-projecting points from one view to another using the derived extrinsics and depth, we observe strong visual correspondence and accurate overlap masks, consistent with the 0.066\% median relative depth error reported in \cref{fig:data_geometric_validation}.

\section{Aerial Image Retrieval}
\label{sec:supp-retrieval}

\subsection{Training on AirZoo}
\myparagraph{Training setup.}
We fine-tune the pre-trained MegaLoc~\cite{berton2025megaloc} on AirZoo with sequence-balanced batching and weighted contrastive supervision. Training is conducted on two NVIDIA RTX 4090 GPUs. During fine-tuning, only the last backbone block and the feature aggregation network are updated, while the remaining backbone layers are frozen. The batch size is set to 85, and each mini-batch contains at most one pair from the same sequence, which increases negative diversity and avoids sequence-specific shortcuts. We train for 10 epochs using AdamW with an initial learning rate of $1 \times 10^{-5}$. The learning rate follows cosine annealing with a 0.1-epoch linear warmup.

\myparagraph{Weighted contrastive loss.}
Standard InfoNCE assumes all positive pairs have equal reliability, which is suitable for one-to-one exact matches. However, UAV--satellite pairs in AirZoo are only partially overlapped. Treating these pairs with uniform positive weight can bias optimization and reduce stability. Therefore, following~\cite{ji2025game4loc}, we use overlap ratio $\text{IOU}_{qr^+}$ as confidence supervision in a weighted InfoNCE objective.

Given a batch of $N$ query-reference pairs, where $F_q$ denotes the query feature and $F_R=\{F_{r_i}\}_{i=1}^{N}$ denotes reference features in the same batch, the weighted loss is:
\begin{equation}
    \begin{aligned}
        \mathcal{L}_\text{weighted-InfoNCE}
        &= \alpha_q \mathcal{L}_\text{InfoNCE}
        + (1-\alpha_q)\mathcal{L}_\text{uniform-InfoNCE}, \\
        \mathcal{L}_\text{InfoNCE}
        &= -\log \frac{\exp(s_{q,r^+}/\tau)}
        {\sum_{i=1}^{N}\exp(s_{q,r_i}/\tau)}, \\
        \mathcal{L}_\text{uniform-InfoNCE}
        &= -\frac{1}{N}\sum_{i=1}^{N}
        \log \frac{\exp(s_{q,r_i}/\tau)}
        {\sum_{j=1}^{N}\exp(s_{q,r_j}/\tau)},
    \end{aligned}
\label{eq:weighted_infonce}
\end{equation}

where $s_{q,r}=F_q\!\cdot\!F_r$, $r^+$ denotes the positive (or semi-positive) reference, and $\tau$ is a learnable temperature. The confidence weight $\alpha_q$ is defined by:
\begin{equation}
    \alpha_q = \sigma(k, \text{IOU}_{qr^+}) = \frac{1}{1 + \exp(-k \times \text{IOU}_{qr^+})},
\label{eq:weight_alpha}
\end{equation}
where $k$ controls the sigmoid curvature and is set to $5$ in all experiments.  More details can be found in Game4Loc~\cite{ji2025game4loc}.

\begin{figure}[!t]
\centering
\includegraphics[width=\textwidth]{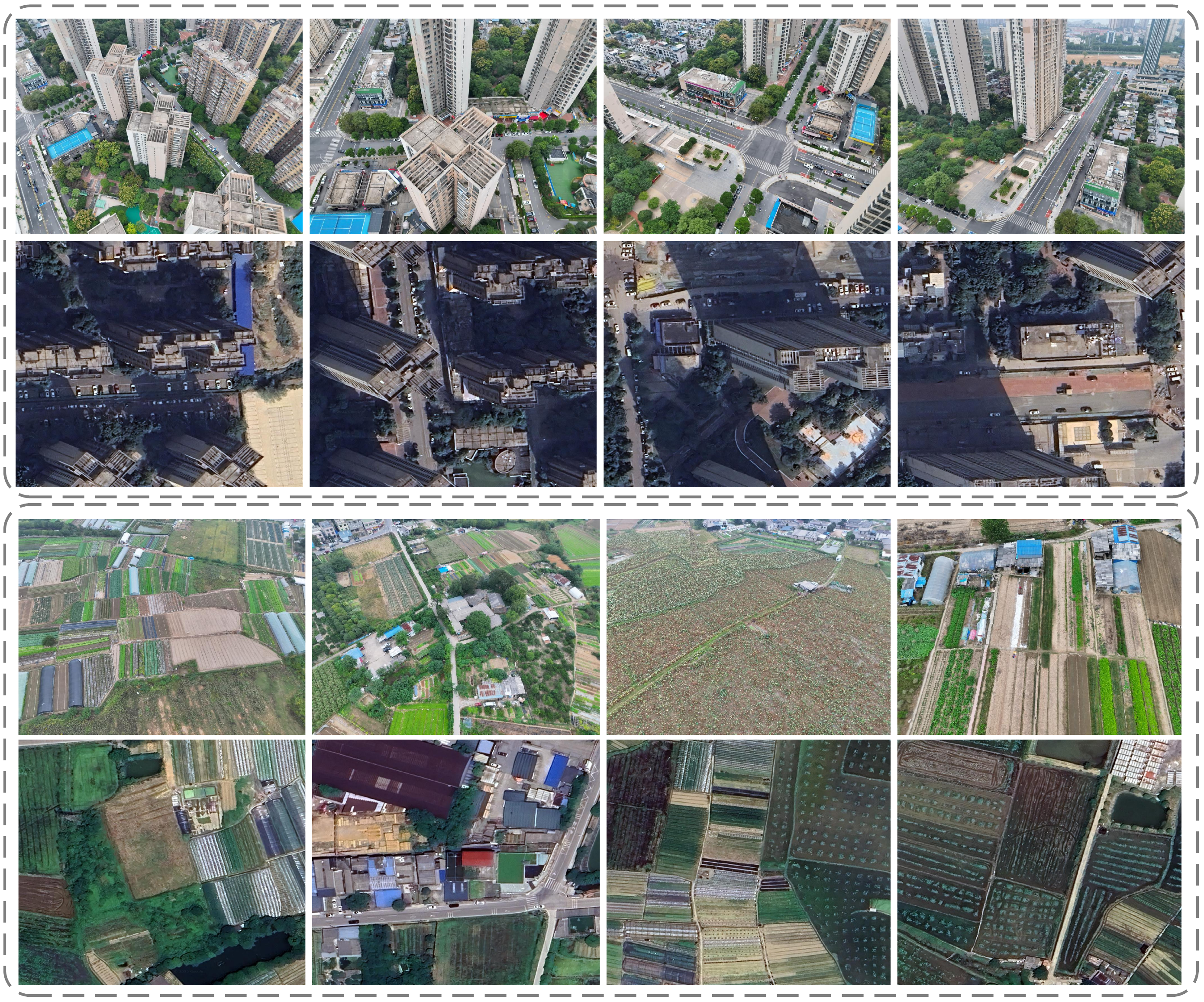}
\caption{\textbf{Qualitative examples from the AirZoo-Real geo-localization subset.} 
The top and bottom rows represent \textit{Urban} (building-dense) and \textit{Rural} (farmland-dominated) scenarios, respectively.
The first column shows real UAV query images, while the second column provides the corresponding ground-truth satellite tiles obtained via our geometric projection pipeline.}
\label{fig:sup_vis_re}
\end{figure}

\myparagraph{Sequence-aware shuffle.}
To maintain diversity throughout training, we apply an epoch-wise shuffle pipeline: (1) randomly shuffle all pairs into a pool; (2) form mini-batches by sequentially sampling from the pool; (3) add a pair only if its sequence label does not already appear in the current mini-batch; and (4) start a new mini-batch once the size reaches 85. This strategy further reduces sequence leakage and improves generalization.

\subsection{Evaluation Benchmark}
\myparagraph{AirZoo-Real (geo-localization subset).}
For retrieval, we use the geo-localization subset of AirZoo-Real, comprising 351 oblique UAV queries over two distinct survey regions in Changsha, each spanning approximately $1.5\,\text{km}^2$.
The query images were captured by an RTK-equipped UAV from diverse oblique viewpoints and low-altitude trajectories.
The resulting dataset comprises two representative settings: an \textit{Urban} scene (161 images) characterized by dense building structures, and a \textit{Rural} scene (190 images) primarily composed of agricultural landscapes.

For reference imagery, we use Zoom-20 satellite maps retrieved from Google Maps.
We build a multi-scale reference gallery by partitioning the high-resolution source maps into tiles of $1024\times1024$ and $512\times512$ pixels.
A $50\%$ overlap ratio is maintained during tiling to ensure spatial continuity and mitigate potential target truncation at tile boundaries, thereby enabling more reliable cross-view retrieval and alignment.

\begin{figure}[!t]
    \centering
    \includegraphics[width=0.9\textwidth]{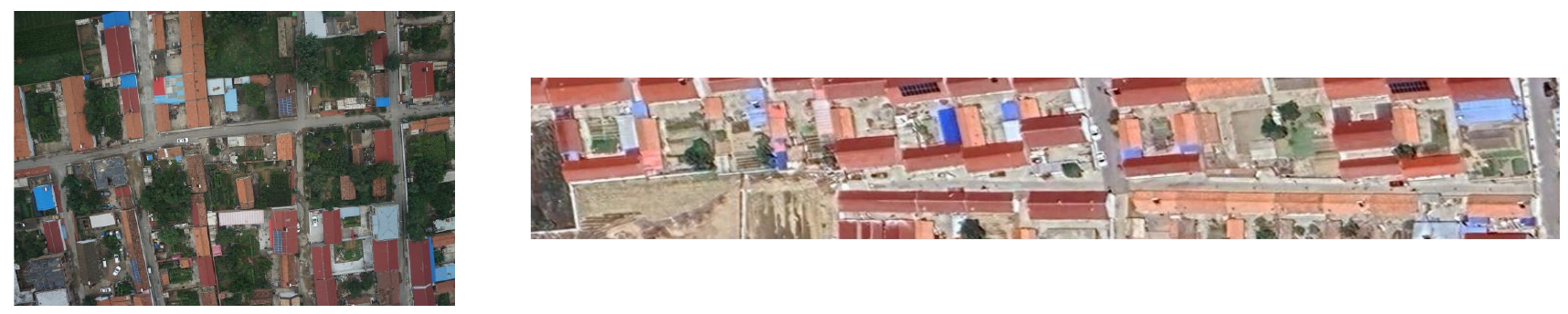}
    \caption{\textbf{Visualization of Scene 7 in UAV-VisLoc.} The left panel shows the query image, and the right panel shows the corresponding satellite map patch, whose scale ratio differs from that in the training data.}
    \label{fig:scene7_vis}
\end{figure}

\myparagraph{Ground truth construction.}
To establish reliable correspondence between UAV queries and the satellite gallery, we adopt a ray-casting protocol to determine the geographical footprint of each query.
Specifically, we sample four anchor vertices near the image corners to represent the camera's field-of-view (FoV):
\[
\mathcal{P}_{uv} = \left\{ (u_i, v_i) \right\}_{i=1}^4 = 
\left\{
\left(\tfrac{W}{16},\tfrac{H}{16}\right),
\left(\tfrac{15W}{16},\tfrac{H}{16}\right),
\left(\tfrac{15W}{16},\tfrac{15H}{16}\right),
\left(\tfrac{W}{16},\tfrac{15H}{16}\right)
\right\}.
\]
Using the calibrated camera intrinsics $\mathbf{K}$, the world-to-camera pose $(\mathbf{R}, \mathbf{t})$, and a reference ground elevation $z_0$ (estimated as the median of valid DSM heights within the region), each pixel is back-projected to a 3D ray and intersected with the ground plane $z=z_0$. The intersection points $\mathbf{X}_i$ in the world coordinate system are computed as:
\[
\mathbf{d}_i^c = \frac{\mathbf{K}^{-1}[u_i,v_i,1]^\top}{\|\mathbf{K}^{-1}[u_i,v_i,1]^\top\|}, \quad
\mathbf{C} = -\mathbf{R}^\top\mathbf{t}, \quad
\mathbf{d}_i^w = \mathbf{R}^\top\mathbf{d}_i^c, \quad
\mathbf{X}_i = \mathbf{C} + \frac{z_0 - C_z}{d_{i,z}^w} \mathbf{d}_i^w.
\]
We then transform $(X_{i,x}, X_{i,y})$ from the local projection system (EPSG:4547) to the global geodetic system (WGS84).
A reference tile is assigned a positive label for a given query if the footprint of the query overlaps with the tile's spatial extent, defined by at least one projected point residing within the tile boundaries $\mathcal{B}_{\text{tile}}$:
\[
\exists i \in \{1,2,3,4\} \quad \text{s.t.} \quad (\text{lon}_i, \text{lat}_i) \in \mathcal{B}_{\text{tile}}.
\]
Qualitative examples of the resulting query-satellite pairs are shown in Fig.~\ref{fig:sup_vis_re}.

\begin{table}[t]
\centering
\scriptsize
\caption{\textbf{Scene-wise geo-localization results on the UAV-VisLoc dataset.} Two scenes are shown side-by-side in each block, and the last block reports Scene 11 with the overall average. Improvements of MegaLoc (Ours) over MegaLoc are shown in gray, and best values are highlighted in \textbf{bold}.}
\setlength{\tabcolsep}{4pt}
\renewcommand{\arraystretch}{1.1}
\resizebox{\linewidth}{!}{%
\begin{tabular}{lcccccccc}
\toprule
\multicolumn{1}{c}{} & \multicolumn{4}{c}{\textbf{Scene 1}} & \multicolumn{4}{c}{\textbf{Scene 2}} \\
\cmidrule(lr){2-5}\cmidrule(lr){6-9}
\textbf{Method} & \textbf{R@1} & \textbf{R@3} & \textbf{R@5} & \textbf{AP@5} & \textbf{R@1} & \textbf{R@3} & \textbf{R@5} & \textbf{AP@5} \\
AnyLoc~\cite{keetha2023anyloc} & 0.86 & 2.20 & 2.57 & 0.61 & 0.56 & 1.21 & 1.77 & 0.43 \\
SALAD~\cite{izquierdo2024optimal} & 28.27 & 38.92 & 45.65 & 16.08 & 22.97 & 32.12 & 37.44 & 13.78 \\
Game4Loc~\cite{ji2025game4loc} & 21.79 & 35.01 & 42.72 & 14.61 & 22.60 & 33.89 & 38.75 & 14.14 \\
MegaLoc~\cite{berton2025megaloc} & 35.99 & 49.33 & 56.30 & 22.11 & 20.82 & 28.48 & 33.15 & 13.37 \\
\rowcolor{yellow!25}
MegaLoc (Ours) & \textbf{40.64} {\color{gray}(+\textbf{4.65})} & \textbf{58.26} {\color{gray}(+\textbf{8.93})} & \textbf{63.28} {\color{gray}(+\textbf{6.98})} & \textbf{25.97} {\color{gray}(+\textbf{3.86})} & \textbf{27.82} {\color{gray}(+\textbf{7.00})} & \textbf{39.87} {\color{gray}(+\textbf{11.39})} & \textbf{45.28} {\color{gray}(+\textbf{12.13})} & \textbf{18.28} {\color{gray}(+\textbf{4.91})} \\
\midrule
\multicolumn{1}{c}{} & \multicolumn{4}{c}{\textbf{Scene 3}} & \multicolumn{4}{c}{\textbf{Scene 4}} \\
\cmidrule(lr){2-5}\cmidrule(lr){6-9}
\textbf{Method} & \textbf{R@1} & \textbf{R@3} & \textbf{R@5} & \textbf{AP@5} & \textbf{R@1} & \textbf{R@3} & \textbf{R@5} & \textbf{AP@5} \\
AnyLoc~\cite{keetha2023anyloc} & 0.00 & 0.00 & 0.00 & 0.00 & 0.14 & 0.14 & 0.14 & 0.08 \\
SALAD~\cite{izquierdo2024optimal} & 16.41 & 24.22 & 27.86 & 8.28 & 29.40 & 40.38 & 44.85 & 16.45 \\
Game4Loc~\cite{ji2025game4loc} & 17.32 & 28.26 & 33.46 & 10.26 & 14.50 & 24.66 & 29.81 & 9.40 \\
MegaLoc~\cite{berton2025megaloc} & 40.10 & 54.43 & 59.38 & 23.75 & 34.28 & 46.75 & 50.95 & 20.46 \\
\rowcolor{yellow!25}
MegaLoc (Ours) & \textbf{41.80} {\color{gray}(+\textbf{1.70})} & \textbf{57.03} {\color{gray}(+\textbf{2.60})} & \textbf{64.06} {\color{gray}(+\textbf{4.68})} & \textbf{26.64} {\color{gray}(+\textbf{2.89})} & \textbf{36.04} {\color{gray}(+\textbf{1.76})} & \textbf{58.81} {\color{gray}(+\textbf{12.06})} & \textbf{64.50} {\color{gray}(+\textbf{13.55})} & \textbf{24.50} {\color{gray}(+\textbf{4.04})} \\
\midrule
\multicolumn{1}{c}{} & \multicolumn{4}{c}{\textbf{Scene 5}} & \multicolumn{4}{c}{\textbf{Scene 6}} \\
\cmidrule(lr){2-5}\cmidrule(lr){6-9}
\textbf{Method} & \textbf{R@1} & \textbf{R@3} & \textbf{R@5} & \textbf{AP@5} & \textbf{R@1} & \textbf{R@3} & \textbf{R@5} & \textbf{AP@5} \\
AnyLoc~\cite{keetha2023anyloc} & 1.06 & 2.96 & 3.81 & 1.31 & 5.33 & 10.65 & 13.02 & 3.37 \\
SALAD~\cite{izquierdo2024optimal} & 17.76 & 39.11 & 50.32 & 15.31 & 30.77 & 41.42 & 47.34 & 17.22 \\
Game4Loc~\cite{ji2025game4loc} & 23.89 & 40.59 & 49.05 & 17.08 & 32.84 & 45.56 & 52.37 & 19.17 \\
MegaLoc~\cite{berton2025megaloc} & 31.50 & 51.16 & 60.47 & 22.54 & 28.99 & 42.90 & 54.14 & 21.36 \\
\rowcolor{yellow!25}
MegaLoc (Ours) & \textbf{42.92} {\color{gray}(+\textbf{11.42})} & \textbf{64.69} {\color{gray}(+\textbf{13.53})} & \textbf{71.04} {\color{gray}(+\textbf{10.57})} & \textbf{28.50} {\color{gray}(+\textbf{5.96})} & \textbf{35.80} {\color{gray}(+\textbf{6.81})} & \textbf{50.00} {\color{gray}(+\textbf{7.10})} & \textbf{55.62} {\color{gray}(+\textbf{1.48})} & \textbf{22.31} {\color{gray}(+\textbf{0.95})} \\
\midrule
\multicolumn{1}{c}{} & \multicolumn{4}{c}{\textbf{Scene 7}} & \multicolumn{4}{c}{\textbf{Scene 8}} \\
\cmidrule(lr){2-5}\cmidrule(lr){6-9}
\textbf{Method} & \textbf{R@1} & \textbf{R@3} & \textbf{R@5} & \textbf{AP@5} & \textbf{R@1} & \textbf{R@3} & \textbf{R@5} & \textbf{AP@5} \\
AnyLoc~\cite{keetha2023anyloc} & 19.23 & 46.15 & \textbf{92.31} & 31.54 & 0.00 & 0.48 & 0.97 & 0.19 \\
SALAD~\cite{izquierdo2024optimal} & \textbf{42.31} & 73.08 & \textbf{92.31} & 36.15 & 2.71 & 6.00 & 8.52 & 1.96 \\
Game4Loc~\cite{ji2025game4loc} & 38.46 & 57.69 & 88.46 & 30.77 & 6.29 & 13.46 & 18.39 & 5.13 \\
MegaLoc~\cite{berton2025megaloc} & 38.46 & \textbf{76.92} & 84.62 & \textbf{38.46} & 8.62 & 13.17 & 15.39 & 4.96 \\
\rowcolor{yellow!25}
MegaLoc (Ours) & 26.92 {\color{gray}(--\textbf{11.54})} & 65.38 {\color{gray}(--\textbf{11.54})} & 84.62 {\color{gray}(+\textbf{0.00})} & 24.62 {\color{gray}(--\textbf{13.84})} & \textbf{9.29} {\color{gray}(+\textbf{0.67})} & \textbf{18.10} {\color{gray}(+\textbf{4.93})} & \textbf{23.23} {\color{gray}(+\textbf{7.84})} & \textbf{6.97} {\color{gray}(+\textbf{2.01})} \\
\midrule
\multicolumn{1}{c}{} & \multicolumn{4}{c}{\textbf{Scene 9}} & \multicolumn{4}{c}{\textbf{Scene 10}} \\
\cmidrule(lr){2-5}\cmidrule(lr){6-9}
\textbf{Method} & \textbf{R@1} & \textbf{R@3} & \textbf{R@5} & \textbf{AP@5} & \textbf{R@1} & \textbf{R@3} & \textbf{R@5} & \textbf{AP@5} \\
AnyLoc~\cite{keetha2023anyloc} & 0.45 & 0.90 & 1.35 & 0.36 & 0.69 & 2.08 & 3.47 & 0.97 \\
SALAD~\cite{izquierdo2024optimal} & 4.95 & 11.26 & 15.77 & 4.10 & 14.58 & 34.72 & 42.36 & 14.41 \\
Game4Loc~\cite{ji2025game4loc} & 10.81 & 21.62 & 29.95 & 8.47 & 6.94 & 17.36 & 25.69 & 7.92 \\
MegaLoc~\cite{berton2025megaloc} & 10.81 & 19.37 & 26.58 & 7.84 & \textbf{25.69} & 38.19 & 48.61 & 15.83 \\
\rowcolor{yellow!25}
MegaLoc (Ours) & \textbf{15.90} {\color{gray}(+\textbf{5.09})} & \textbf{27.93} {\color{gray}(+\textbf{8.56})} & \textbf{36.94} {\color{gray}(+\textbf{10.36})} & \textbf{10.90} {\color{gray}(+\textbf{3.06})} & \textbf{25.69} {\color{gray}(+\textbf{0.00})} & \textbf{45.14} {\color{gray}(+\textbf{6.95})} & \textbf{50.69} {\color{gray}(+\textbf{2.08})} & \textbf{19.58} {\color{gray}(+\textbf{3.75})} \\
\midrule
\multicolumn{1}{c}{} & \multicolumn{4}{c}{\textbf{Scene 11}} & \multicolumn{4}{c}{\textbf{Average}} \\
\cmidrule(lr){2-5}\cmidrule(lr){6-9}
\textbf{Method} & \textbf{R@1} & \textbf{R@3} & \textbf{R@5} & \textbf{AP@5} & \textbf{R@1} & \textbf{R@3} & \textbf{R@5} & \textbf{AP@5} \\
AnyLoc~\cite{keetha2023anyloc} & 0.00 & 0.17 & 0.17 & 0.07 & 2.57 & 6.09 & 10.87 & 3.54 \\
SALAD~\cite{izquierdo2024optimal} & 4.41 & 9.83 & 14.58 & 3.56 & 19.50 & 31.91 & 38.82 & 13.39 \\
Game4Loc~\cite{ji2025game4loc} & 10.00 & 20.17 & 27.63 & 8.27 & 18.68 & 30.75 & 39.66 & 13.20 \\
MegaLoc~\cite{berton2025megaloc} & 7.63 & 19.66 & 24.92 & 6.81 & 25.72 & 40.03 & 46.77 & 17.95 \\
\rowcolor{yellow!25}
MegaLoc (Ours) & \textbf{12.20} {\color{gray}(+\textbf{4.57})} & \textbf{24.41} {\color{gray}(+\textbf{4.75})} & \textbf{31.86} {\color{gray}(+\textbf{6.94})} & \textbf{9.32} {\color{gray}(+\textbf{2.51})} & \textbf{28.64} {\color{gray}(+\textbf{2.92})} & \textbf{46.33} {\color{gray}(+\textbf{6.30})} & \textbf{53.74} {\color{gray}(+\textbf{6.96})} & \textbf{19.78} {\color{gray}(+\textbf{1.83})} \\
\bottomrule
\end{tabular}%
}
\label{tab:uav_visloc_scenes}
\end{table}

\begin{table}[t]
\centering
\scriptsize
\caption{\textbf{Scene-wise geo-localization results on the AirZoo-Real.} City and Rural scenes are shown side-by-side, followed by a full-width average block. Improvements of MegaLoc (Ours) over MegaLoc are shown in gray, and best values are highlighted in \textbf{bold}.}
\setlength{\tabcolsep}{4pt}
\renewcommand{\arraystretch}{1.1}
\resizebox{\linewidth}{!}{%
\begin{tabular}{lcccccccc}
\toprule
\multicolumn{1}{c}{} & \multicolumn{4}{c}{\textbf{City}} & \multicolumn{4}{c}{\textbf{Rural}} \\
\cmidrule(lr){2-5}\cmidrule(lr){6-9}
\textbf{Method} & \textbf{R@1} & \textbf{R@3} & \textbf{R@5} & \textbf{AP@5} & \textbf{R@1} & \textbf{R@3} & \textbf{R@5} & \textbf{AP@5} \\
AnyLoc~\cite{keetha2023anyloc} & 0.62 & 1.86 & 2.48 & 0.62 & 8.42 & 16.84 & 18.42 & 5.26 \\
SALAD~\cite{izquierdo2024optimal} & 6.83 & 16.77 & 24.22 & 6.71 & 7.89 & 21.58 & 34.74 & 8.74 \\
Game4Loc~\cite{ji2025game4loc} & \textbf{21.12} & \textbf{36.65} & \textbf{47.83} & \textbf{16.15} & 4.21 & 10.53 & 20.53 & 5.16 \\
MegaLoc~\cite{berton2025megaloc} & 8.70 & 19.88 & 24.84 & 7.08 & 4.21 & 13.16 & 19.47 & 5.58 \\
\rowcolor{yellow!25}
MegaLoc (Ours) & 9.94 {\color{gray}(+\textbf{1.24})} & 22.98 {\color{gray}(+\textbf{3.10})} & 31.06 {\color{gray}(+\textbf{6.22})} & 9.69 {\color{gray}(+\textbf{2.61})} & \textbf{27.37} {\color{gray}(+\textbf{23.16})} & \textbf{58.42} {\color{gray}(+\textbf{45.26})} & \textbf{70.00} {\color{gray}(+\textbf{50.53})} & \textbf{24.74} {\color{gray}(+\textbf{19.16})} \\
\midrule
\multicolumn{1}{c}{} & \multicolumn{8}{c}{\textbf{Average}} \\
\cmidrule(lr){2-9}
\textbf{Method} & \multicolumn{2}{c}{\textbf{R@1}} & \multicolumn{2}{c}{\textbf{R@3}} & \multicolumn{2}{c}{\textbf{R@5}} & \multicolumn{2}{c}{\textbf{AP@5}} \\
AnyLoc~\cite{keetha2023anyloc} & \multicolumn{2}{c}{4.52} & \multicolumn{2}{c}{9.35} & \multicolumn{2}{c}{10.45} & \multicolumn{2}{c}{2.94} \\
SALAD~\cite{izquierdo2024optimal} & \multicolumn{2}{c}{7.36} & \multicolumn{2}{c}{19.17} & \multicolumn{2}{c}{29.48} & \multicolumn{2}{c}{7.72} \\
Game4Loc~\cite{ji2025game4loc} & \multicolumn{2}{c}{12.67} & \multicolumn{2}{c}{23.59} & \multicolumn{2}{c}{34.18} & \multicolumn{2}{c}{10.65} \\
MegaLoc~\cite{berton2025megaloc} & \multicolumn{2}{c}{6.46} & \multicolumn{2}{c}{16.52} & \multicolumn{2}{c}{22.16} & \multicolumn{2}{c}{6.33} \\
\rowcolor{yellow!25}
MegaLoc (Ours) & \multicolumn{2}{c}{\textbf{18.66} {\color{gray}(+\textbf{12.20})}} & \multicolumn{2}{c}{\textbf{40.70} {\color{gray}(+\textbf{24.18})}} & \multicolumn{2}{c}{\textbf{50.53} {\color{gray}(+\textbf{28.38})}} & \multicolumn{2}{c}{\textbf{17.21} {\color{gray}(+\textbf{10.88})}} \\
\bottomrule
\end{tabular}%
}
\label{tab:uav_real_scenes}
\end{table}

\subsection{Results}
\myparagraph{Per-scene evaluation on UAV-VisLoc.}
Table~\ref{tab:uav_visloc_scenes} provides a detailed breakdown of retrieval performance across 11 scenes in the UAV-VisLoc benchmark. Compared with the baseline MegaLoc, our fine-tuned variant, \textit{MegaLoc (Ours)}, achieves substantial improvements in 10 of 11 scenes on R@3, R@5, and AP@5, while maintaining a clear lead in 9 scenes on R@1. Notably, the largest gains are observed in Scene 5 ($+11.42$ on R@1 and $+13.53$ on R@3) and Scene 2 ($+7.00$ on R@1 and $+12.13$ on R@5).

An outlier is Scene 7, where \textit{MegaLoc (Ours)} exhibits a slight performance drop. This scene has a constrained geographic extent and satellite tiles with atypical aspect-ratio distributions that are underrepresented in the training corpus. Moreover, limited database coverage requires aggressive resizing to satisfy our fixed-crop protocol, introducing non-negligible geometric distortion. This out-of-distribution (OOD) condition likely explains the observed drop, as visualized in Fig.~\ref{fig:scene7_vis}.

\myparagraph{Per-scene results on AirZoo-Real.}
Table~\ref{tab:uav_real_scenes} evaluates cross-scene generalization on the two real-flight regions. MegaLoc (Ours) demonstrates consistent superiority over the original model in both City and Rural environments. We observe that Game4Loc~\cite{ji2025game4loc} performs competitively in the City scene, where domain shifts are relatively mild, but degrades significantly in the Rural scenario. In contrast, our model maintains robust performance across both domains. This trend suggests that AirZoo-based fine-tuning not only improves retrieval accuracy but also yields stronger cross-scene generalization, achieving the best results among compared methods under diverse real-world conditions.

\section{Cross-view Matching}
\label{sec:supp-matching}

\subsection{Training on AirZoo}
\myparagraph{Training setup.} We leverage RoMa~\cite{edstedt2024roma} as our baseline method and fine-tune it on a curated mixture of the MegaDepth~\cite{li2018megadepth} and AirZoo datasets. 
The training process is parallelized across 8 NVIDIA RTX 5090 GPUs with a per-GPU batch size of 4, yielding a total effective batch size of $B_{\text{eff}}=32$. 
All input images are resized to a resolution of $560\times560$. 
The network weights are initialized from the publicly available RoMa checkpoint pre-trained on outdoor scenes~\cite{edstedt2024roma}.

To balance the data distribution during training, we employ a weighted sampling strategy with a MegaDepth-to-AirZoo ratio of $80\%:20\%$. 
Specifically, we set the overlap tiers to $[0.01, 0.35]$ for MegaDepth and $[0.3, 0.5]$ for AirZoo. 
Optimization is performed using AdamW with a weight decay of $0.01$. 
To preserve the pre-trained feature representations while allowing for task-specific adaptation, we adopt differential learning rates for the encoder and decoder:
\begin{equation}
    \eta_{\text{enc}}=2\times10^{-7}, \quad \eta_{\text{dec}}=4\times10^{-6}.
\end{equation}
The model is trained for a total of $250\text{k}$ iterations, with a learning rate decay by a factor of $0.1$ applied at the $50\text{k}$ step mark.

\subsection{Evaluation Framework}
\myparagraph{AirZoo-Real benchmark.}
We evaluate the performance of cross-view matching on the AirZoo-Real benchmark, consistent with the setup employed in the aerial retrieval experiments. 
In this task, the reference gallery consists of Digital Orthophoto Maps (DOM) and Digital Surface Models (DSM) reconstructed from high-precision nadir UAV flights. 
The query set comprises authentic oblique UAV images captured across diverse urban and rural landscapes. 
This configuration presents a significant challenge due to the drastic perspective shifts between the nadir reference maps and the oblique query views.

\myparagraph{GT construction.}
The ground-truth translations are derived from high-precision RTK-recorded UAV positions. 
Conversely, since IMU-based orientations are susceptible to cumulative drift over extended flight trajectories, they lack the requisite reliability for a rigorous 6-DoF pose evaluation. 
Consequently, we focus on translation-centric metrics, reporting the median translation error alongside the recall rate at distance thresholds of $5$\,m, $10$\,m, and $20$\,m.

\begin{figure}[ht]
\centering
\includegraphics[width=\textwidth]{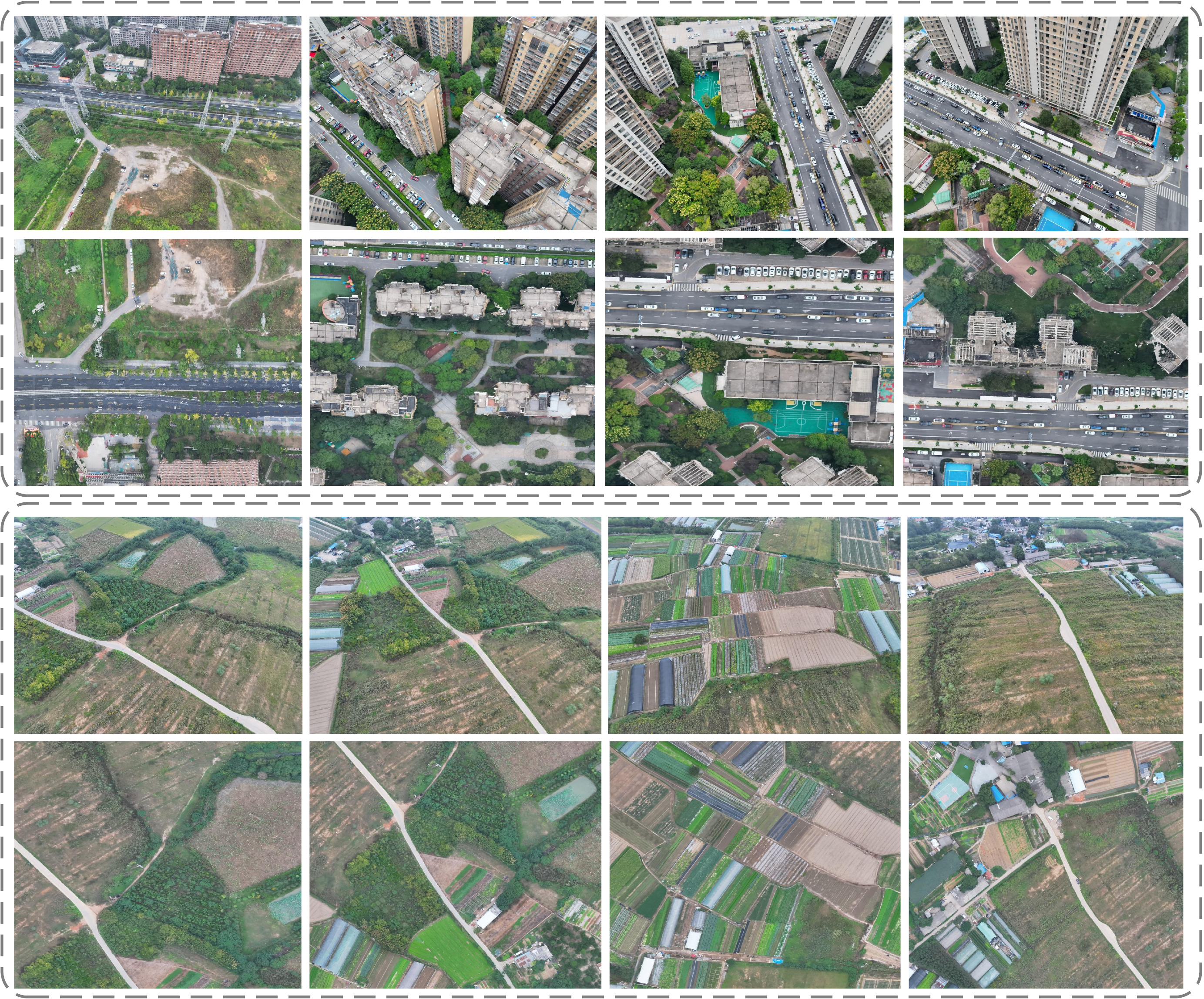}
\caption{\textbf{Visualization of AirZoo-Real geo-localization pairs.} Each pair contains an oblique UAV query and its DOM reference tile. Top and bottom rows show representative City and Rural cases, respectively.}
\label{fig:match_vis_pair}
\end{figure}

For each query image, we leverage the on-board sensor priors (GPS position and IMU orientation) to project the four image corner vertices onto the georeferenced DOM/DSM, thereby extracting a corresponding reference crop for matching. The visualizations of these constructed matching pairs across diverse urban and rural scenarios are presented in Fig.~\ref{fig:match_vis_pair}.

\subsection{Results}
\begin{table}[t]
\centering
\scriptsize
\caption{\textbf{Cross-view matching results on AirZoo-Real.} City and Rural scenes are reported side-by-side. Gray values indicate changes relative to the original RoMA, and best results are highlighted in \textbf{bold}.}
\setlength{\tabcolsep}{3pt}
\renewcommand{\arraystretch}{1.1}
\resizebox{\linewidth}{!}{%
\begin{tabular}{lcccccccc}
\toprule
\multicolumn{1}{c}{} & \multicolumn{4}{c}{\textbf{City (161)}} & \multicolumn{4}{c}{\textbf{Rural (190)}} \\
\cmidrule(lr){2-5}\cmidrule(lr){6-9}
\textbf{Method} & \textbf{Med.} $\downarrow$ & \textbf{@5m} $\uparrow$ & \textbf{@10m} $\uparrow$ & \textbf{@20m} $\uparrow$ & \textbf{Med.} $\downarrow$ & \textbf{@5m} $\uparrow$ & \textbf{@10m} $\uparrow$ & \textbf{@20m} $\uparrow$ \\
\midrule
LoFTR~\cite{sun2021loftr} & 265.25 & 1.86 & 7.45 & 9.94 & 223.14 & 4.21 & 6.32 & 6.32 \\
ELoFTR~\cite{wang2024efficient} & 249.01 & 3.11 & 8.08 & 9.32 & 210.26 & 4.74 & 7.37 & 9.47 \\
DUSt3R~\cite{wang2024dust3r} & 210.33 & 1.24 & 5.59 & 13.04 & 218.81 & 3.68 & 10.00 & 15.26 \\
MASt3R~\cite{murai2025mast3r} & 206.17 & 11.18 & 18.01 & 22.98 & 161.33 & 14.21 & 22.11 & 25.26 \\
RoMA~\cite{edstedt2024roma} & 6.08 & 44.10 & 55.90 & 57.76 & 5.66 & 45.79 & 55.26 & 56.84 \\
\rowcolor{brown!15}
RoMA (GIM)~\cite{xuelun2024gim} &
3.35 {\color{gray}(--2.73)} &
73.30 {\color{gray}(+29.20)} &
81.44 {\color{gray}(+25.54)} &
\textbf{83.85} {\color{gray}(+26.09)} &
3.08 {\color{gray}(--2.58)} &
82.11 {\color{gray}(+36.32)} &
\textbf{86.84} {\color{gray}(+31.58)} &
\textbf{90.00} {\color{gray}(+33.16)} \\
\rowcolor{yellow!25}
RoMA (Ours) &
\textbf{3.07} {\color{gray}(--3.01)} &
\textbf{76.40} {\color{gray}(+32.30)} &
\textbf{83.85} {\color{gray}(+27.95)} &
\textbf{83.85} {\color{gray}(+26.09)} &
\textbf{2.99} {\color{gray}(--2.67)} &
\textbf{83.16} {\color{gray}(+37.37)} &
\textbf{86.84} {\color{gray}(+31.58)} &
\textbf{90.00} {\color{gray}(+33.16)} \\
\bottomrule
\end{tabular}%
}
\label{tab:matching_results}
\end{table}

The quantitative evaluation on the AirZoo-Real geo-localization subset is summarized in Table~\ref{tab:matching_results}. Our fine-tuned RoMa model demonstrates substantial and consistent performance gains over the original RoMa and other compared baselines across both urban and rural environments. As shown in Table~\ref{tab:matching_results}, conventional matching methods such as LoFTR~\cite{sun2021loftr} and the more recent MASt3R~\cite{murai2025mast3r} struggle significantly with the drastic perspective shifts and domain gaps inherent in this benchmark, yielding high median errors and low recall rates. In contrast, our fine-tuning strategy leverages the diverse cross-view geometry of AirZoo to substantially bridge this gap. Specifically, compared to the original RoMa, RoMa (Ours) reduces the median translation error by nearly half: from $6.08$\,m to $3.07$\,m in \textit{City} and from $5.66$\,m to $2.99$\,m in \textit{Rural}.

\begin{figure}[ht]
\centering
\includegraphics[width=\textwidth]{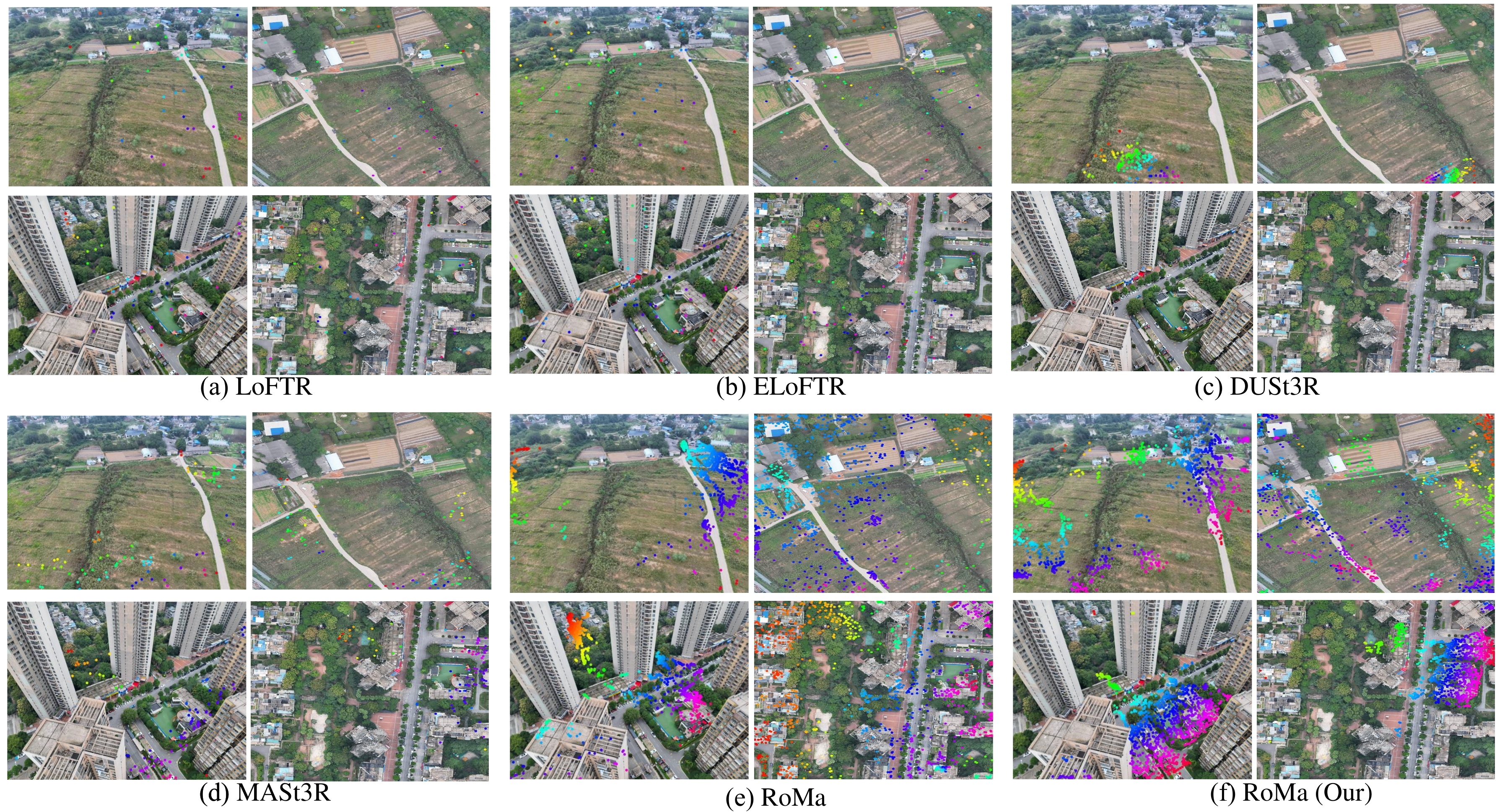}
\caption{\textbf{Qualitative comparison of matching performance on the AirZoo-Real geo-localization subset.} 
We visualize the correspondences between oblique UAV query images and their corresponding DOM reference tiles. 
Compared with the baselines (a--e), which often yield sparse or locally clustered matches under extreme viewpoint changes, our fine-tuned RoMa (f) produces significantly denser and more spatially uniform inlier correspondences.}
\label{fig:match_vis_points}
\end{figure}

\section{Multi-view 3D Reconstruction}
\label{sec:supp-reconstruction}

\subsection{Training on AirZoo}
\myparagraph{Training setup.}
We select 16 trajectories from Brazil, the USA, and New Zealand as our validation/test set, and use the remaining 361 trajectories as the training set to fine-tune our baseline models, VGGT~\cite{wang2025vggt} and DA3~\cite{lin2025depth}, on 8 NVIDIA A100 Tensor Core GPUs. During training, we first randomly determine both the overlap ratio and the number of images in each sampled sequence, with the overlap ratio constrained to 50\%--75\%. Following VGGT and DA3, we set the sequence length to a random value in the range 2--24 for VGGT and 2--10 for DA3. We then randomly choose a start point from the full image stream that satisfies the required overlap and sequence length, and generate the corresponding image sequence. This protocol exposes the models to diverse temporal contexts and viewpoint baselines, improving robustness to varying motion patterns and scene layouts. The randomized overlap further regularizes training by preventing overfitting to a fixed sampling strategy, while maintaining sufficient inter-frame redundancy for stable multi-view geometry learning.

\subsection{Evaluation Benchmark}
\myparagraph{AirZoo-Real (reconstruction subset).}
For reconstruction, we use a separate real-flight subset of AirZoo-Real with 9,430 images across four representative regions in Changsha: a school, a driving school, a substation, and a plaza. The substation scene is characterized by dense vegetation, whereas the remaining regions are predominantly urban. Each scene was captured during three time periods (6:00--8:00, 12:00--14:00, and 18:00--20:00) with approximately consistent viewpoints across time slots; the school and driving school scenes additionally include low-light data from 22:00--24:00. Flights were conducted at approximately 160\,m altitude with pitch angles between 30$^\circ$ and 45$^\circ$, and RTK positioning was used throughout to ensure accurate extrinsics. Depth supervision was obtained by reconstructing 3D models via oblique photogrammetry and rendering depth maps at the capture viewpoints. Further acquisition details are provided in \cref{sec:experiments}.

\begin{figure}[!t]
\centering
\includegraphics[width=\textwidth]{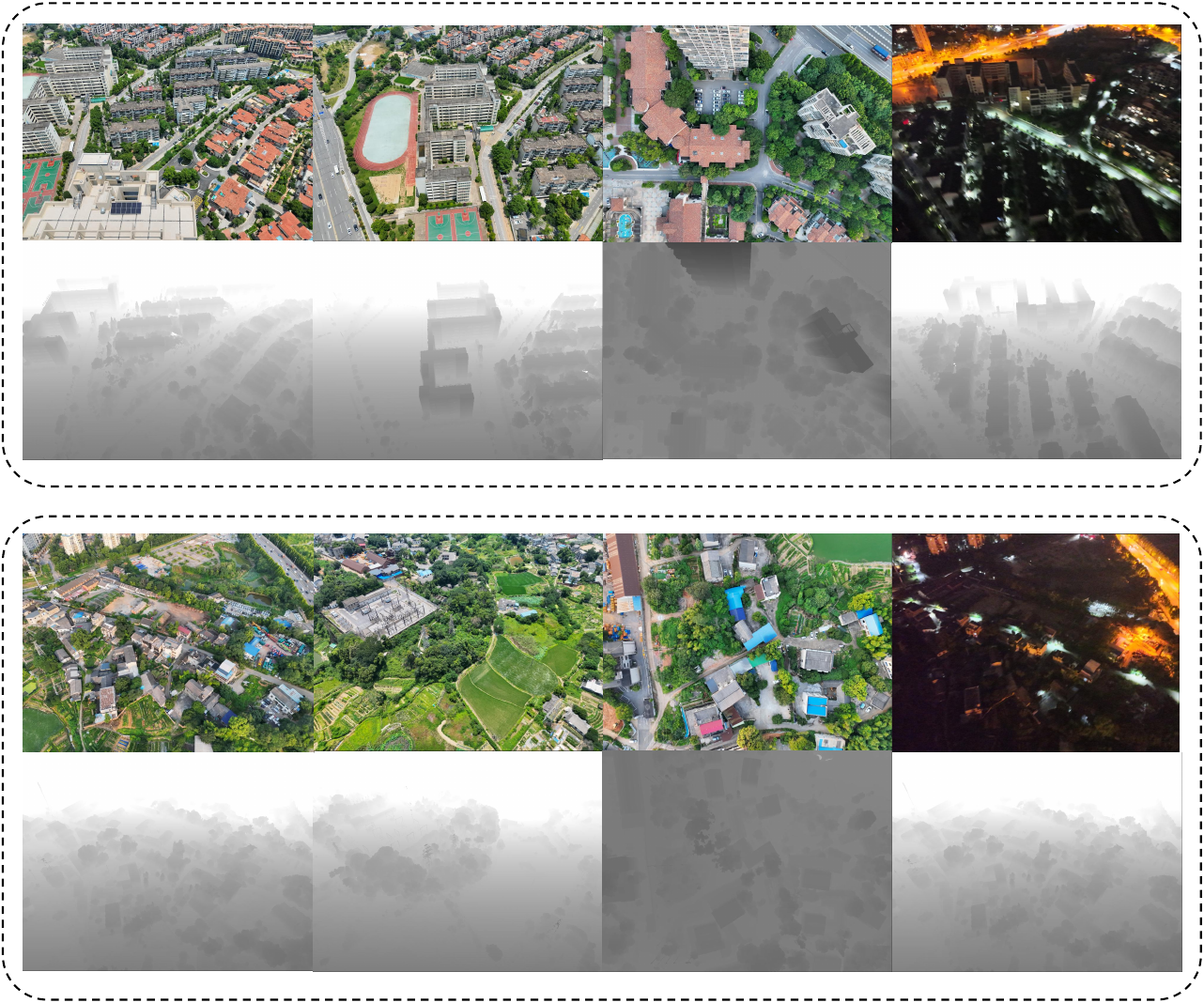}
\caption{\textbf{Visualization of the AirZoo-Real reconstruction subset.} Representative oblique UAV queries (left) and the corresponding rendered depth supervision (right) from the four real-flight scenes.}
\label{fig:supp_recon_vis}
\end{figure}

\myparagraph{Evaluation Metrics.}

Let $G$ denote the ground-truth point set and $R$ denote the reconstructed point set to be evaluated. We use $\mathrm{dist}(R\!\to\!G)$ to measure reconstruction accuracy and $\mathrm{dist}(G\!\to\!R)$ to measure reconstruction completeness. The Chamfer Distance (CD) is defined as the average of these two distances. Based on these distances, we define the precision and recall of the reconstruction $R$ under a distance threshold $d$:
\begin{itemize}
  \item Precision:
  $$
  \frac{1}{|R|}\sum_{i}\big[\mathrm{dist}(R_i \to G) < d\big]
  $$
  \item Recall:
  $$
  \frac{1}{|G|}\sum_{i}\big[\mathrm{dist}(G_i \to R) < d\big]
  $$
\end{itemize}
To jointly reflect precision and recall performance, we report the F1 score, computed as
$$
F1 = \frac{2 \times \text{Precision} \times \text{Recall}}{\text{Precision} + \text{Recall}}.
$$

\begin{figure}[!ht]
    \centering
    \includegraphics[width=0.9\textwidth]{figs/supp_recon_result.pdf}
    \caption{\textbf{Qualitative reconstruction results on AirZoo-Real, UAVScenes, and UrbanScene3D.} Left: query image. Right: reconstruction result.}
    \label{fig:supp_recon}
\end{figure}

\subsection{Results}
Quantitative results on AirZoo-Real, UAVScenes, UrbanScene3D, and AirZoo-Test are reported in \cref{tab:uav-recon}. \Cref{fig:supp_recon} provides qualitative comparisons. VGGT-SLAM~\cite{maggio2025vggt}, although initialized from the VGGT checkpoint and refined with bundle adjustment, still misses portions of scene geometry on UrbanScene3D~\cite{lin2022capturing}. CUT3R~\cite{wang2025continuous} shows clear scale inconsistencies on AirZoo-Real, likely due to the absence of aerial-domain training. For DA3, our fine-tuned model produces reconstructions with fewer visible artifacts and stronger geometric consistency than the baseline.

\subsection{Failure Case.}
In our failure case analysis, we observed that because the AirZoo dataset does not contain image sequences captured along circular trajectories around individual buildings, the fine-tuned model—despite outperforming the baseline on standard forward flight paths—fails to correctly infer the 3D structural relationships of a building when processing aerial sequences that orbit a single structure. Lacking the training signal to associate the building's appearance from multiple surround viewpoints, the model is unable to establish a coherent geometric understanding of the scene. This deficiency directly leads to significant drift in the estimated camera poses or to severe geometric distortion and texture misalignment in the reconstructed building, ultimately resulting in failure cases.

\end{document}